%% file: icml2021.tex
\documentclass{article}
\usepackage{booktabs} 

\usepackage[utf8]{inputenc} 
\usepackage[T1]{fontenc}    
\usepackage{hyperref}       
\usepackage{url}            
\usepackage{amsmath}
\usepackage{amsfonts}       
\usepackage{nicefrac}       
\usepackage{microtype}      
\usepackage{graphicx}
\usepackage{xcolor}
\usepackage{subcaption}
\usepackage{wrapfig}
\usepackage{relsize}
\usepackage{lipsum} 
\usepackage{multicol}
\usepackage{caption}
\usepackage{makecell}
\usepackage[inline]{enumitem}
\setlist[enumerate]{topsep=0pt,itemsep=0.1mm,partopsep=1ex,parsep=1ex}

\usepackage{pifont}
\newcommand{\cmark}{\ding{51}}%
\newcommand{\xmark}{\ding{55}}%

\usepackage[accepted]{icml2021}

\icmltitlerunning{Vector Quantized Models for Planning}

\definecolor{customcolor}{rgb}{0.82, 0.41, 0.12}

\newcommand{\z}{\mathbf{z}}
\newcommand{\x}{\mathbf{x}}

\input{math_commands.tex}

\graphicspath{{./figures/}}

\begin{document}

\twocolumn[
\icmltitle{Vector Quantized Models for Planning}



\icmlsetsymbol{equal}{*}


\begin{icmlauthorlist}
\icmlauthor{Sherjil Ozair}{equal,dm,mila}
\icmlauthor{Yazhe Li}{equal,dm}
\icmlauthor{Ali Razavi}{dm}
\icmlauthor{Ioannis Antonoglou}{dm}
\icmlauthor{A\"aron van den Oord}{dm}
\icmlauthor{Oriol Vinyals}{dm}

\end{icmlauthorlist}

\icmlaffiliation{dm}{DeepMind, London, United Kingdom}
\icmlaffiliation{mila}{Mila, University of Montreal}

\icmlcorrespondingauthor{Sherjil Ozair}{sherjilozair@deepmind.com}
\icmlcorrespondingauthor{Yazhe Li}{yazhe@deepmind.com}

\icmlkeywords{Machine Learning, ICML}

\vskip 0.3in
]



\printAffiliationsAndNotice{\icmlEqualContribution} 

\input{abstract.tex}

\input{introduction-oriol.tex}

\input{related_work.tex}

\input{background.tex}

\input{mcts.tex}

\input{experiments.tex}

\input{discussion.tex}
\bibliography{icml2021}
\bibliographystyle{icml2021}

\twocolumn[
\newpage
]

\appendix
\input{appendix.tex}

\end{document}

%% file: math_commands.tex
\usepackage{amsmath,amsfonts,bm}

\def\eqref#1{equation~\ref{#1}}

\def\1{\bm{1}}

\DeclareMathAlphabet{\mathsfit}{\encodingdefault}{\sfdefault}{m}{sl}
\SetMathAlphabet{\mathsfit}{bold}{\encodingdefault}{\sfdefault}{bx}{n}

\DeclareMathOperator*{\argmax}{arg\,max}

%% file: abstract.tex
\begin{abstract}
Recent developments in the field of model-based RL have proven successful in a range of environments, especially ones where planning is essential. However, such successes have been limited to deterministic fully-observed environments. We present a new approach that handles stochastic and partially-observable environments. Our key insight is to use discrete autoencoders to capture the multiple possible effects of an action in a stochastic environment. We use a stochastic variant of \emph{Monte Carlo tree search} to plan over both the agent's actions and the discrete latent variables representing the environment's response. Our approach significantly outperforms an offline version of MuZero on a stochastic interpretation of chess where the opponent is considered part of the environment. We also show that our approach scales to \emph{DeepMind Lab}, a first-person 3D environment with large visual observations and partial observability.
\end{abstract}

%% file: introduction-oriol.tex
\section{Introduction}

Making predictions about the world may be a necessary ingredient towards building intelligent agents, as humans use these predictions to devise and enact plans to reach complex goals \citep{lake2017building}. However, in the field of reinforcement learning (RL), a tension still exists between model-based and model-free RL. Model-based RL and planning have been key ingredients in many successes such as games like chess \citep{shannon1950xxii, alphazero}, Go \citep{alphago, alphagozero}, and Poker \citep{moravvcik2017deepstack, brown2017libratus}. However, their applicability to richer environments with larger action and state spaces remains limited due to some of the key assumptions made in such approaches. Other notable results have not used any form of model or planning, such as playing complex video games Dota 2 \citep{openai2019dota} and StarCraft II \citep{Vinyals2019}, or robotics \citep{DBLP:journals/corr/abs-1808-00177}.

\begin{figure}[htbp]
     \centering
     \begin{subfigure}[h]{0.9\columnwidth}
         \centering
          \includegraphics[width=\columnwidth]{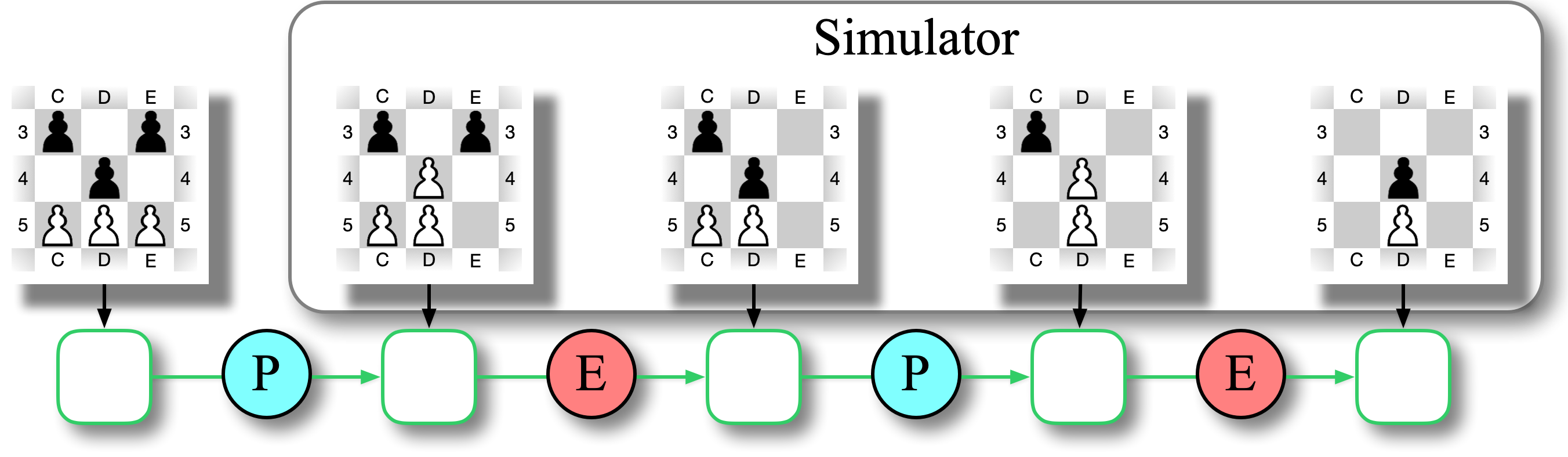}
         \caption{AlphaZero}
         \label{fig:alphazero_path}
     \end{subfigure}
     \begin{subfigure}[h]{0.9\columnwidth}
         \centering
         \includegraphics[width=\columnwidth]{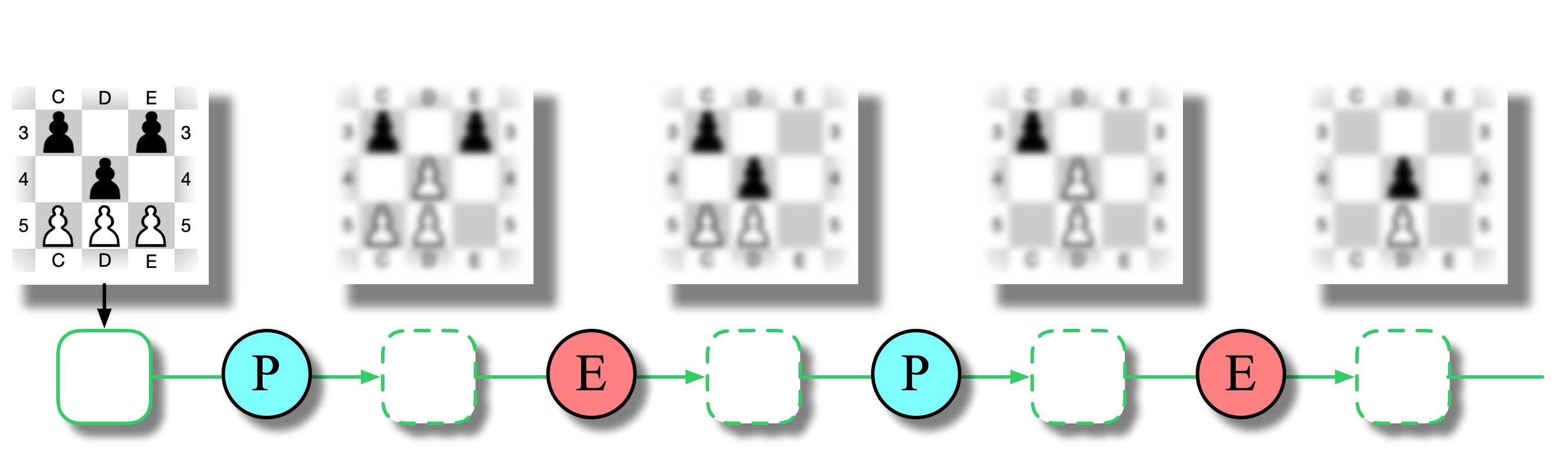}
         \caption{Two-player MuZero}
         \label{fig:two_player_muzero_path}
     \end{subfigure}     
     \begin{subfigure}[h]{0.9\columnwidth}
         \centering
         \includegraphics[width=\columnwidth]{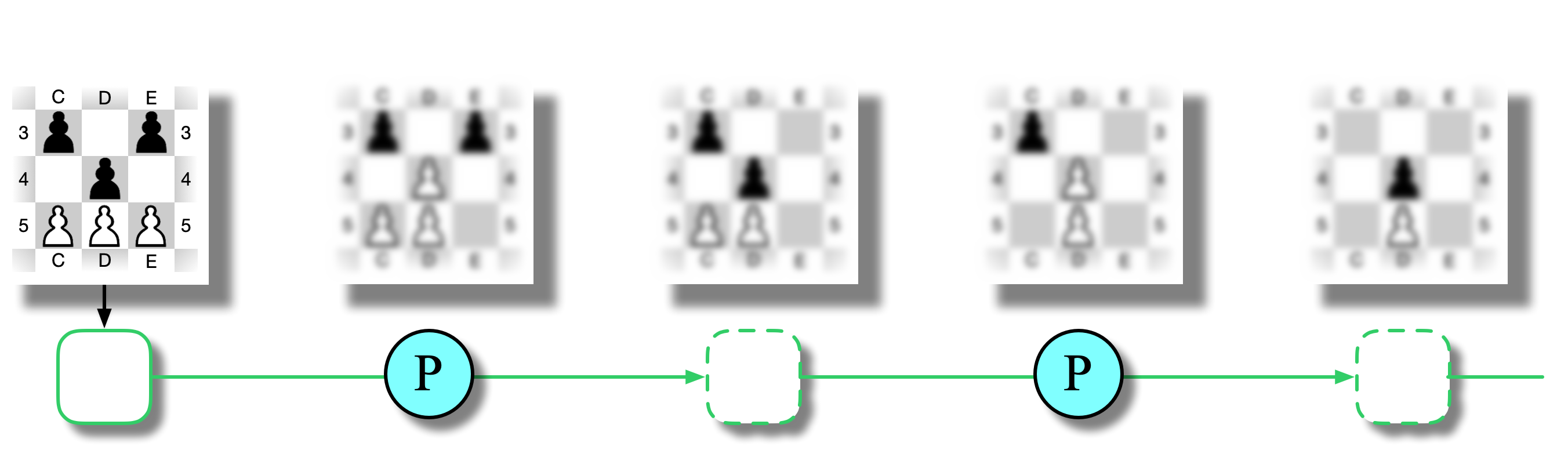}
         \caption{Single-player MuZero}
         \label{fig:single_player_muzero_path}
     \end{subfigure}
     
     \begin{subfigure}[h]{0.9\columnwidth}
         \centering
         \includegraphics[width=\columnwidth]{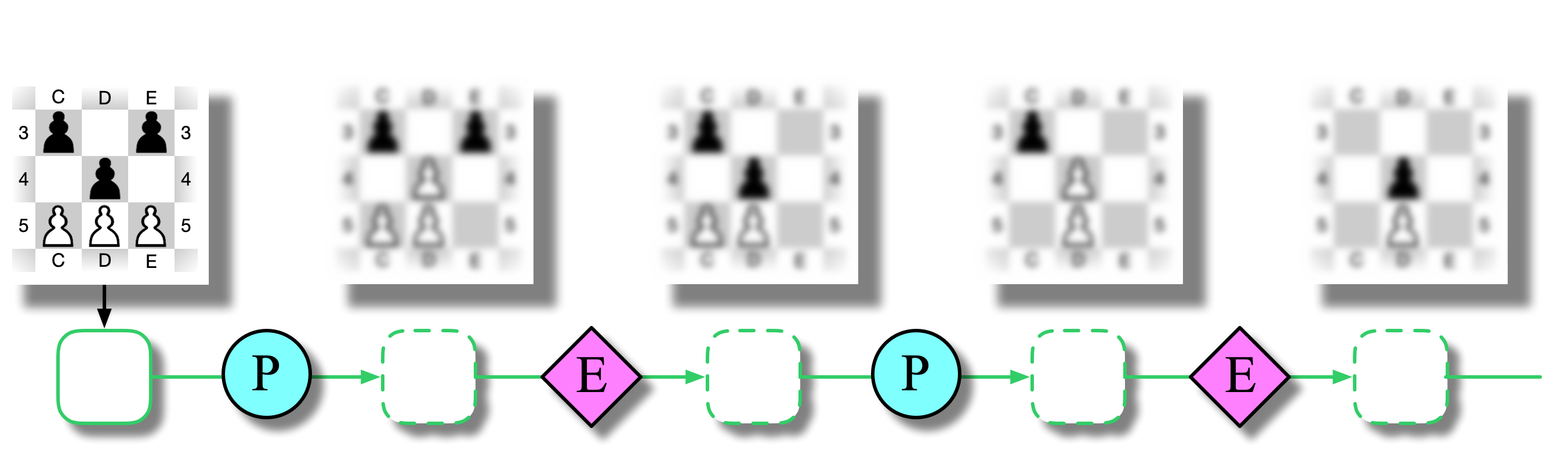}
         \caption{VQHybrid}
         \label{fig:vq_hybrid}
     \end{subfigure}
     \begin{subfigure}[h]{0.9\columnwidth}
         \centering
         \includegraphics[width=\columnwidth]{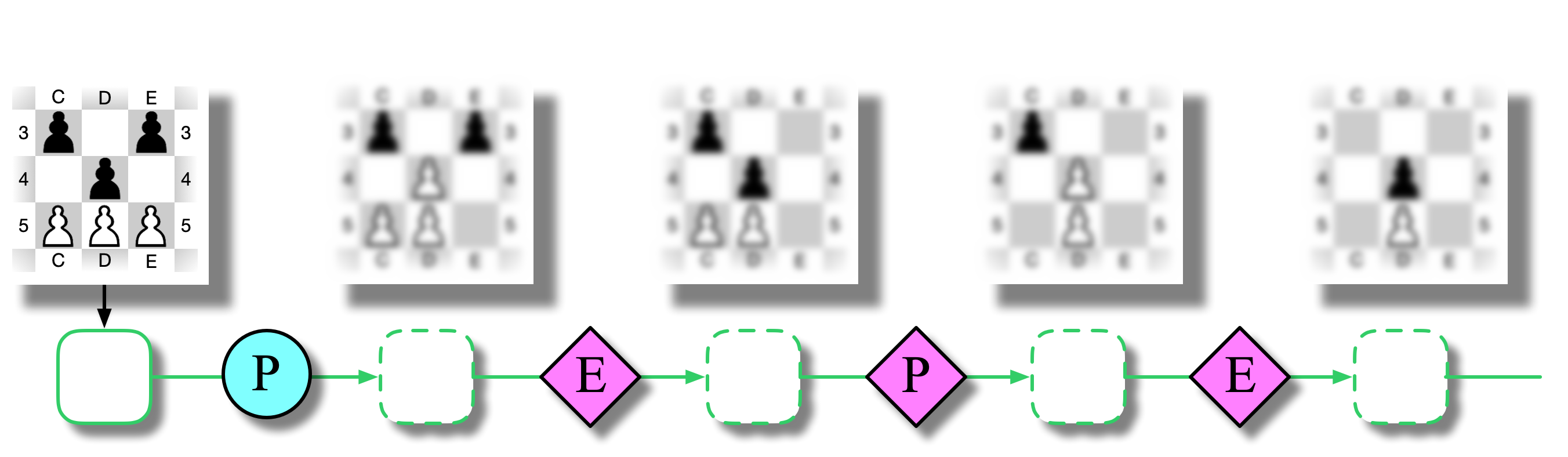}
         \caption{VQPure}
         \label{fig:vq_pure}
     \end{subfigure}
     \begin{subfigure}[h]{0.9\columnwidth}
         \centering
         \includegraphics[width=\columnwidth]{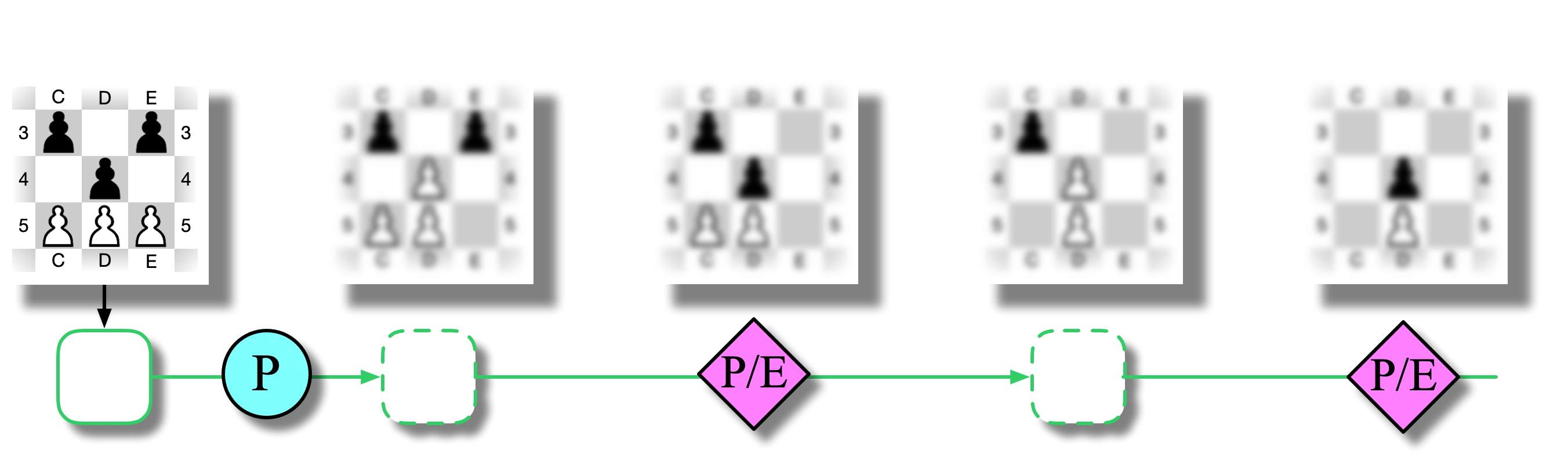}
         \caption{VQJumpy}
         \label{fig:vq_jumpy}
     \end{subfigure}
     \caption{\textbf{Comparison of model-based planning formulations.} \textbf{(\subref{fig:alphazero_path})} AlphaZero plans with both player and opponent actions and the groundtruth states with the help of a simulator; \textbf{(\subref{fig:two_player_muzero_path})} Two-player version of MuZero plans with both player and opponent actions; \textbf{(\subref{fig:single_player_muzero_path})} Single-player MuZero plans only with the player actions; \textbf{(\subref{fig:vq_hybrid})} VQHybrid plans with player actions and discrete latent variables; \textbf{(\subref{fig:vq_pure})} VQPure plans with the player action for the first step and discrete latent variables thereafter; \textbf{(\subref{fig:vq_jumpy})} VQJumpy plans with discrete latent variables that expand for more than a single agent step. \\
     \textbf{Notations:} The rounded squares denote states. Circles are actions in real action space. Diamonds are discrete latent variables. \\ P denotes the player's action. E denotes the environment's action.
     }
     \label{fig:planning_path}
\end{figure}

In this work we are motivated by widening the applicability of model-based planning by devising a solution which removes some of the key assumptions made by the MuZero algorithm \citep{schrittwieser2019mastering}. \autoref{table:method_compare} and \autoref{fig:planning_path} summarize the key features of model-based planning algorithms discussed in this paper. MuZero lifts the crucial requirement of having access to a perfect simulator of the environment dynamics found in previous model-based planning approaches \citep{alphazero, anthony2017thinking}. In many cases such a simulator is not available (eg., weather forecasting), is expensive (eg., scientific modeling), or is cumbersome to run (e.g. for complex games such as Dota 2 or StarCraft II).

\begin{table}[b]
\vskip -0.3in

\caption{Key features of different planning algorithms.}
\label{table:method_compare}

\vskip 0.1in
\begin{center}
\resizebox{\columnwidth}{!}{
\begin{tabular}{cccccc}
\toprule
     Method & \makecell{Learned \\ Model} & \makecell{Agent \\ Perspective} & Stochastic & \makecell{Abstract \\ Actions} & \makecell{Temporal \\ Abstraction} \\
\midrule
AlphaZero & \xmark & \xmark & \xmark  & \xmark & \xmark \\ \midrule
      \makecell{Two-player \\ MuZero} & \cmark & \xmark & \xmark  & \xmark & \xmark \\ \midrule
      \makecell{Single-player \\ MuZero} & \cmark & \cmark & \xmark  & \xmark & \xmark \\ \midrule
     VQHybrid & \cmark & \cmark  & \cmark & \xmark & \xmark \\ \midrule
     VQPure & \cmark & \cmark  & \cmark & \cmark & \xmark \\ \midrule
     VQJumpy & \cmark & \cmark & \cmark  & \cmark & \cmark \\
\bottomrule
\end{tabular}
}
\end{center}
\vskip -0.1in
\end{table}

However, MuZero still makes a few limiting assumptions. It assumes the environment to be deterministic, limiting which environments can be used. It assumes full access to the state, also limiting which environments can be used. The search and planning is over future agent(s) actions, which could be millions in environments with complex action spaces. The search occurs at every agent-environment interaction step, which may be too fine grained and wasteful.

Largely inspired by both MuZero and the recent successes of VQVAEs \citep{van2017neural, razavi2019generating} and large language models \citep{radford2019language, brown2020language}, we devise VQ models for planning, which in principle can remove most of these assumptions.

Our approach uses a state VQVAE and a transition model. The state VQVAE encodes future observations into discrete latent variables. This allows the use of \textit{Monte Carlo tree search} (MCTS, \cite{coulom2006efficient}) for planning not only over future actions, but also over future observations, thus allowing planning in stochastic or partially-observed environments.

We also propose a ``pure'' version of our model which encodes both future observations and actions into discrete latent variables, which would allow planning entirely in discrete latent variables. These discrete latent variables can be designed to have cardinality and time granularity independent of the actions, thus enabling planning in large action spaces over longer time horizons. 

To demonstrate that the proposed solution works well in practice, we devise two evaluation frameworks using the game of chess. The first is the classic \emph{two-player} chess framework where agents get to observe and plan over both their own and their opponent's actions. This is the framework used by techniques like MuZero and its previous iterations. In this setting, all four assumptions above perfectly hold, i.e. action and time granularity is already at the right level of abstraction, there is no stochasticity, and agents observe the full board. Then we remove the ability to enumerate opponent actions, and make the opponent's move part of the environment dynamics. That is agents can only observe their own states and actions. This makes the environment stochastic, since the transition is not a deterministic function but depends on an unknown opponent's action from a potentially stochastic policy. We refer to this framework as \emph{single-player} chess.

We show that MuZero's performance drops catastrophically on \emph{single-player} chess compared to \emph{two-player} chess, demonstrating that MuZero depends on access to data from the opponent's perspective. Our approach which only uses player perspective data at training and playing time performs as well on \emph{single-player} chess as MuZero does on \emph{two-player} chess. This suggests that our approach is a promising way to generalize MuZero-style planning to partially-observable and stochastic environments.

To investigate how well our approach can scale, we also evaluate on DeepMind Lab \citep{beattie2016deepmind}, which has a complex observation and action space with partial observability and stochasticity, showing that VQ planning models capture the uncertainty and offer the best results among all planning models we compare against.

%% file: related_work.tex
\section{Related Work}

\paragraph{Models for Planning}
\citet{oh2015actionconditional, chiappa2017recurrent,kaiser2019model} use video prediction models as environment simulators. However, such models are not feasible for planning since they require observation reconstruction which would make planning prohibitively slow. \citet{hasselt2019use} argue that experience replay can be regarded as a non-parametric model and that Dyna-based methods are unlikely to outperform model-free methods. \citet{schrittwieser2019mastering, oh2017value} learn an implicit deterministic sequence model by predicting future reward, value and policy from current states and future actions. However, these models are in principle limited to deterministic or weakly stochastic environments such as Atari \citep{machado2018revisiting}.

Stochastic models promise to capture uncertainty. 
PILCO \citep{ICML2011Deisenroth} used Gaussian processes for transition model and achieves remarkable sample efficiency by capturing model uncertainty. However, it is not scalable to high dimensional state spaces. \citet{depeweg2016learning} model uncertainty of the transition function with Bayesian neural networks (BNNs). \citet{kurutach2018model, chua2018deep} use model ensembles to capture epistemic uncertainty that arise from scarcity of data. Variational autoencoders (VAE, \citet{kingma2013auto, rezende2014stochastic}) have fuelled a range of stochastic models for RL. \citet{moerland2017learning} builds models for RL with conditional VAE \citep{sohn2015learning}. \citet{buesing2018learning} investigate stochastic state-space models. 
\citet{ha2018recurrent} train a VAE to compress the observation into continuous latent variables and use an RNN to serve as the predictive model.
\citet{hafner2018learning, hafner2019dream} learn a full forward model using VAE framework. They incorporate multi-step prediction (``latent overshooting'') to minimize compounding errors, which changes the optimization objective, while our approach uses data likelihood as the only objective. \citet{hafner2020mastering} propose a discrete autoencoders model and learn it with straight-through gradients. This is perhaps the most similar to our approach. However, the model is used to generate synthetic data and not used for planning. Lastly, \citet{rezende2020causally} study environment models from a causal perspective. They propose adding stochastic nodes using \emph{backdoors} \citep{pearl2016causal}. This approach requires the backdoor variable to be observed and recorded during data generation. Our approach doesn't alter the data generating process, therefore works in offline RL setting.

\paragraph{Model-based policies}
Models can be used in different ways to materialize a policy. \citet{oh2015actionconditional,kaiser2019model} use environment models in the Dyna \citep{Sutton1998} framework, which proposes to learn a policy with model-free algorithms using synthetic experiences generated by models. However, the accumulated error in synthesized data could hurt performance compared to an online agent. This loss of performance has been studied by \citet{hasselt2019use}. \citet{ha2018recurrent,hafner2018learning} do policy improvement through black-box optimization such as CMA-ES, which is compatible with continuous latent variable models. \citet{henaff2017model} extends policy optimization to discrete action space. Following AlphaGo \citep{alphago} and AlphaZero \citep{alphazero}, \citet{schrittwieser2019mastering} prove that MCTS is scalable and effective for policy improvement in model-based learning. Our approach is a generalization of MuZero that is able to incorporate stochasticity and abstract away planning from agent actions. Continuous action spaces and MCTS  have also been combined with some success, e.g. \cite{couetoux2011continuous} and \cite{yee2016monte}. However, our choice of discrete latent space makes it possible to leverage all the recent advances made in MCTS. In a specific multi-agent setup, where the focus is to find policies that are less exploitable, models can be used with counterfactual regret minimization \citep{DBLP:journals/corr/MoravcikSBLMBDW17} or fictitious play \citep{DBLP:journals/corr/HeinrichS16} to derive Nash equilibrium strategy.

\paragraph{Offline RL}

While our model-based approach is applicable generally, we evaluate it in the \emph{offline RL} setting. Previous model-based offline RL approaches \citep{argenson2020model, yu2020mopo, kidambi2020morel} have focused on continuous control problems \citep{tassa2020dmcontrol, gulcehre2020rl}. Our work focuses on environments with large observation spaces and complex strategies which require planning such as chess and DeepMind Lab \citep{beattie2016deepmind}. 


%% file: background.tex
\section{Background}

\paragraph{Vector-Quantized Variational AutoEncoders} 
\label{sec:vqvae}
(VQVAE, \citet{van2017neural}) make use of vector quantization (VQ) to learn discrete latent variables in a variational autoencoder. VQVAE comprises of neural network encoder and decoder, a vector quantization layer, and a reconstruction loss function. The encoder takes as input the data sample $\x$, and outputs vector $\mathbf{z}_{u} = f(\x)$. The vector quantization layer maintains a set of embeddings $\{\mathbf{e}_k\}_{k=1}^{K}$. It outputs an index $c$ and the corresponding embedding $\mathbf{e}_c$, which is closest to the input vector $\mathbf{z}_{u}$ in Euclidean distance.
The decoder neural network uses the embedding $\mathbf{e}_c$ as its input to produce reconstructions $\hat{\x}$. The full loss is $\mathcal{L}^{t} = \mathcal{L}^{r}(\hat{\x}, \x) + \beta \|\z_u - sg(\mathbf{e}_c)\|^2$, where $sg(\cdot)$ is the stop gradient function. The second term is the commitment loss used to regularize the encoder to output vectors close to the embeddings so that error due to quantization is minimized. The embeddings are updated to the exponential moving average of the minibatch average of the unquantized vectors assigned to each latent code. In the backwards pass, the quantization layer is treated as an identity function, referred to as straight-through gradient estimation \citep{bengio2013estimating}. For more details, see \citet{van2017neural}. 








\paragraph{Monte Carlo Tree Search}
\label{sec:mcts}
(MCTS, \citet{coulom2006efficient}) is a tree search method for estimating the optimal action given access to a simulator, typically used in two-player games. MCTS builds a search tree by recursively expanding the tree and assessing the value of the leaf node using Monte Carlo (MC) simulation. Values of leaf nodes are used to estimate the Q-values of all the actions in the root node.

The policy for child node selection during expansion is crucial to the performance of MCTS. The most popular method for this is UCT (stands for ``Upper Confidence Bounds applied to trees'', \citet{kocsis2006bandit}), which is based on Upper Confidence Bound (UCB, \citet{auer2002finite}). UCT suggests that this problem can be seen as a Bandit problem where the optimal solution is to combine the value estimation with its uncertainty.

AlphaGo \citep{44806} combined MCTS with neural networks by using them for value and policy estimations. The benefits of this approach are twofold: value estimation no longer incurs expensive Monte Carlo simulations, allowing for shallow searches to be effective, and the policy network serves as context for the tree expansion and limits the branching factor.

At each search iteration, the MCTS algorithm used in AlphaGo consists of 3 steps: selection, expansion and value backup. During selection stage, MCTS descends the search tree from the root node by picking the action that maximizes the following upper confidence bound:
\begin{equation}
    \argmax_a \left[Q(s, a) + P(a|s) U(s, a)\right], \label{eq:mcts_action}
\end{equation} 
where
\begin{equation}
    U(s, a) = \frac{\sqrt{N(s)}}{1 +N(s, a)} [c_1 + \log (\frac{N(s) + c_2 + 1}{c_2}) ],
\end{equation} and $N(s, a)$ is the visit counts of taking action $a$ at state $s$, $N(s) = \sum_b N(s, b)$ is the number of times $s$ has been visited, $c_1$ and $c_2$ are constants that control the influence of the policy $P(a|s)$ relative to the value $Q(s, a)$.

Following the action selection, the search tree receives the next state. If the next state doesn't already exist in the search tree, a new leaf node is added and this results in an expansion of the tree. The value of the new leaf node is evaluated with the learned value function. Finally, the estimated value is backed up to update MCTS’s value statistics of the nodes along the descending path:
\begin{equation}
    Q^{t+1}_{tree}(s, a) = \frac{Q^{t}_{tree}(s, a) N^{t}(s,a) + Q(s, a)}{N^{t}(s,a) + 1}.
\end{equation} 
Here $Q^{t}_{tree}(s,a)$ is the action value estimated by the tree search at iteration $t$; $N^{t}(s,a)$ is the visit count at iteration $t$.

\paragraph{MuZero}
\label{sec:muzero}
\citet{schrittwieser2019mastering} introduce further advances in MCTS, where a sequential model is learned from trajectories $\{s_0, a_0, \dots, s_{T-1}, a_{T-1}, s_T\}$. At each timestep, the model uses the trajectory to formulate a \emph{planning path} at timestep $t$: $s_t, a_{t}, \dots, a_{t+M-1}, a_{t+M}$ with $s_t$ being the root state. To simplify the notation, we omit the subscript and use superscript for indexing actions on the planning path. So the same sequence is written as $s, a^0, \dots, a^{M-1}, a^{M}$. Given the starting root state $s$ and a sequence of actions $a^{0:m}$, the model outputs a hidden state $h^{m}$ and predicts action policy $\pi^{m}$, value $v^{m}$ and reward $r^{m}$. The training objective is as follows:
\begin{align}
    \frac{1}{M} \sum_{m=1}^M [ &{L}^{\pi}(a^m, \pi(h^m)) + \alpha \mathcal{L}^v(v_{target}^m, v(h^m)) \nonumber \\
    & + \beta  \mathcal{L}^{r}(r_{env}^m, r(h^m))] \label{eq:muzero_loss}
\end{align}
where $h^m$ is the hidden state on the planning path. ${L}^{\pi}(a^m, \pi(h^m))$ is the cross entropy loss between the action and learned parametric policy $\pi$. $\mathcal{L}^v$ is the value loss function, $v_{target}$ is the value target and $v$ is the value prediction. $\mathcal{L}^r$ is the reward loss function, $r_{env}$ is the environment reward and $r$ is the reward prediction. $\alpha$ and $\beta$ are the weights.

During search, $\pi$ is used as the prior for action selection; $r$ gives the reward instead of using reward from the simulator; $v$ estimates the value of the leaf state rather than using Monte Carlo rollouts.

Comparing to AlphaZero, MuZero model eliminates the need of a simulator to generate the groundtruth state along the planning path. In two player games, MuZero's planning path interleaves the player's action and opponent's action. Whereas in single player version, only player's actions are seen by the model.



%% file: mcts.tex
\section{Model-based Planning with VQVAEs}
\label{sec:approach}

Our approach uses a state VQVAE model and a transition model. We refer to the full model as VQ Model (VQM), and the resulting agent when combined with MCTS as VQM-MCTS.

We first describe the components of VQM in detail. Then, we explain how the model is used with MCTS.

\subsection{VQ Model}

\begin{figure}[!ht]
    \begin{center}
    \begin{subfigure}[h]{\columnwidth}
         \centering
          \includegraphics[width=0.85\textwidth]{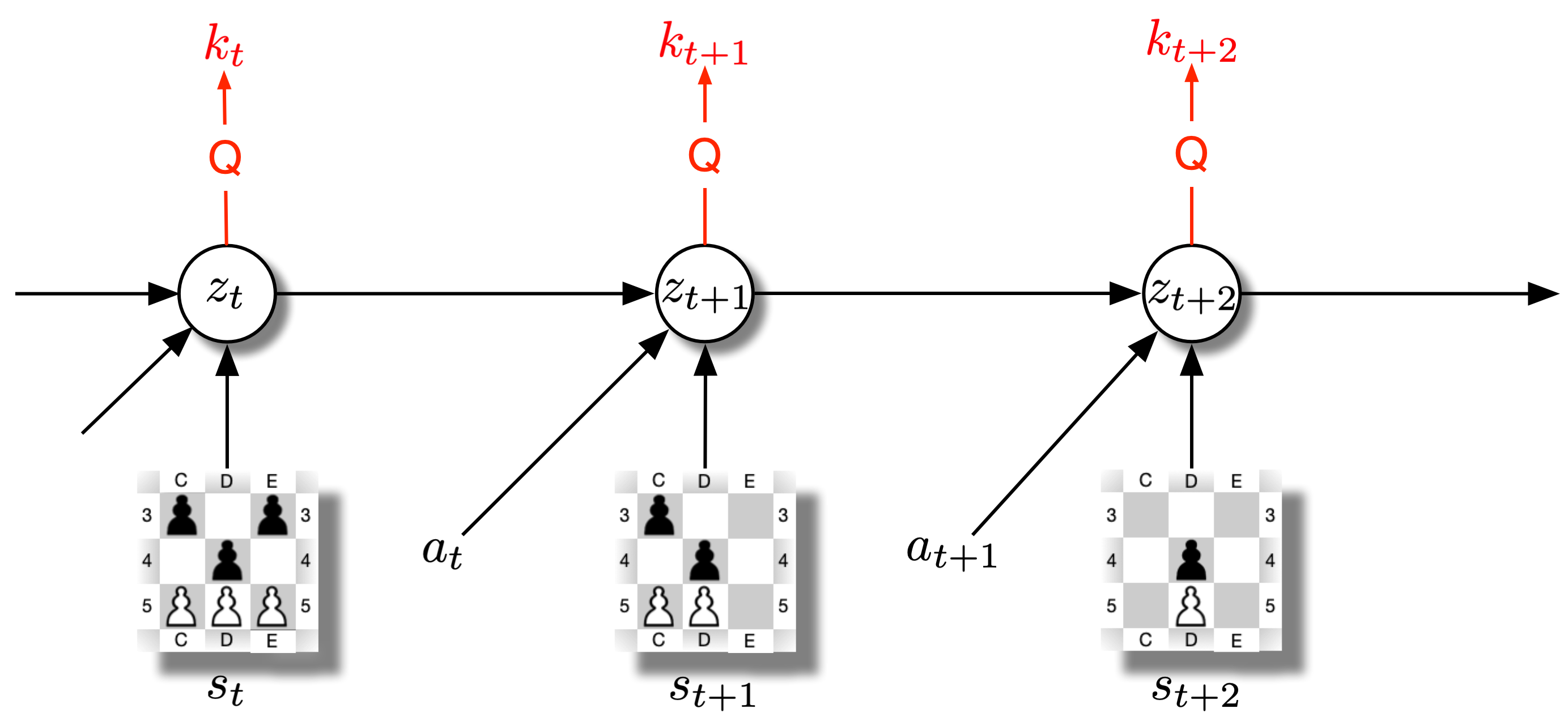}
         \caption{The state VQVAE encodes a sequence of observations $s$ and actions $a$ into discrete latent variables $k$.}
         \vspace{1em}
        \label{fig:state-vqvae}
    \end{subfigure}
    \begin{subfigure}[h]{\columnwidth}
         \centering
          \includegraphics[width=0.85\textwidth]{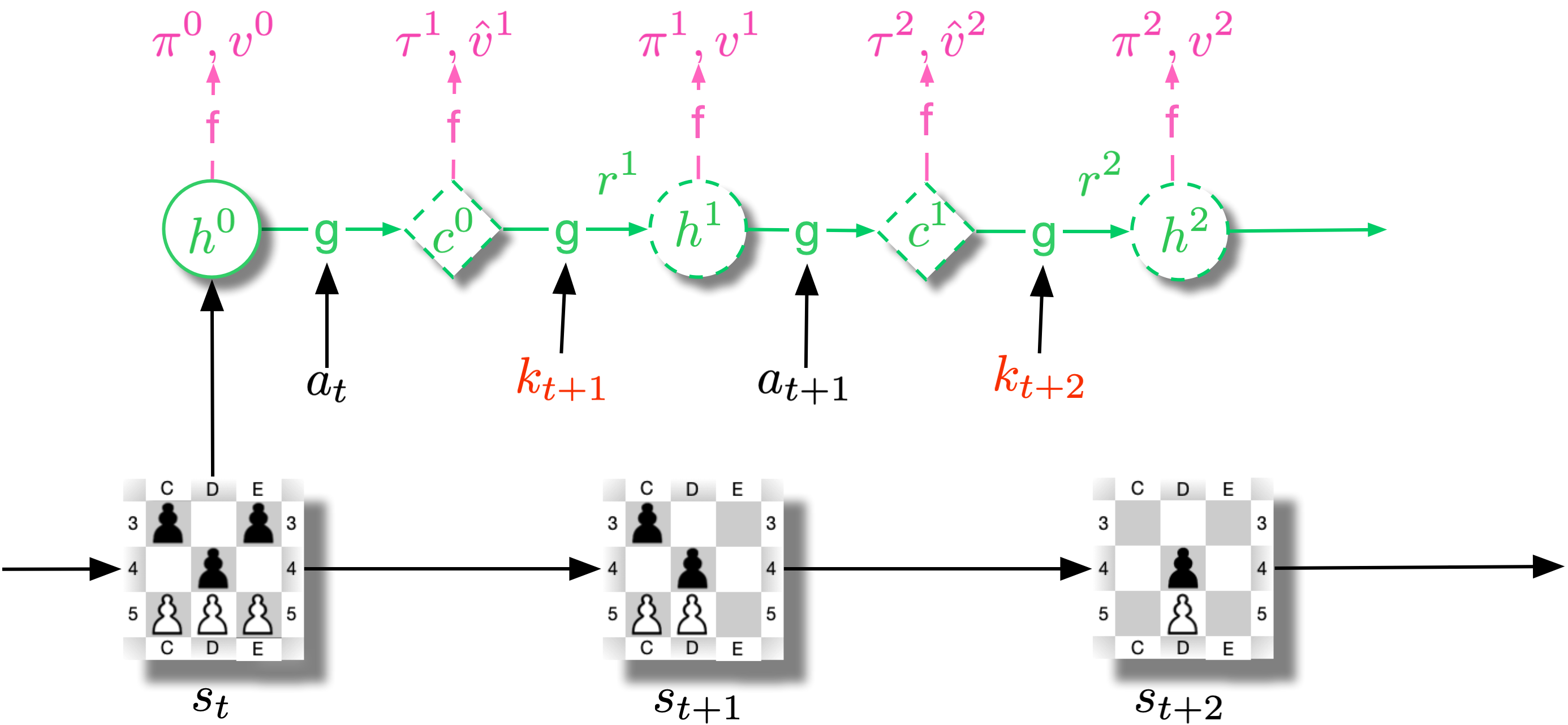}
         \caption{An autoregressive transition model outputs a policy $\pi$ over the actions, a policy $\tau$ over the discrete latent codes and a value function $v$.}
         \label{fig:transition-model}
     \end{subfigure}
     \begin{subfigure}[h]{\columnwidth}
         \centering
          \includegraphics[width=0.7\textwidth]{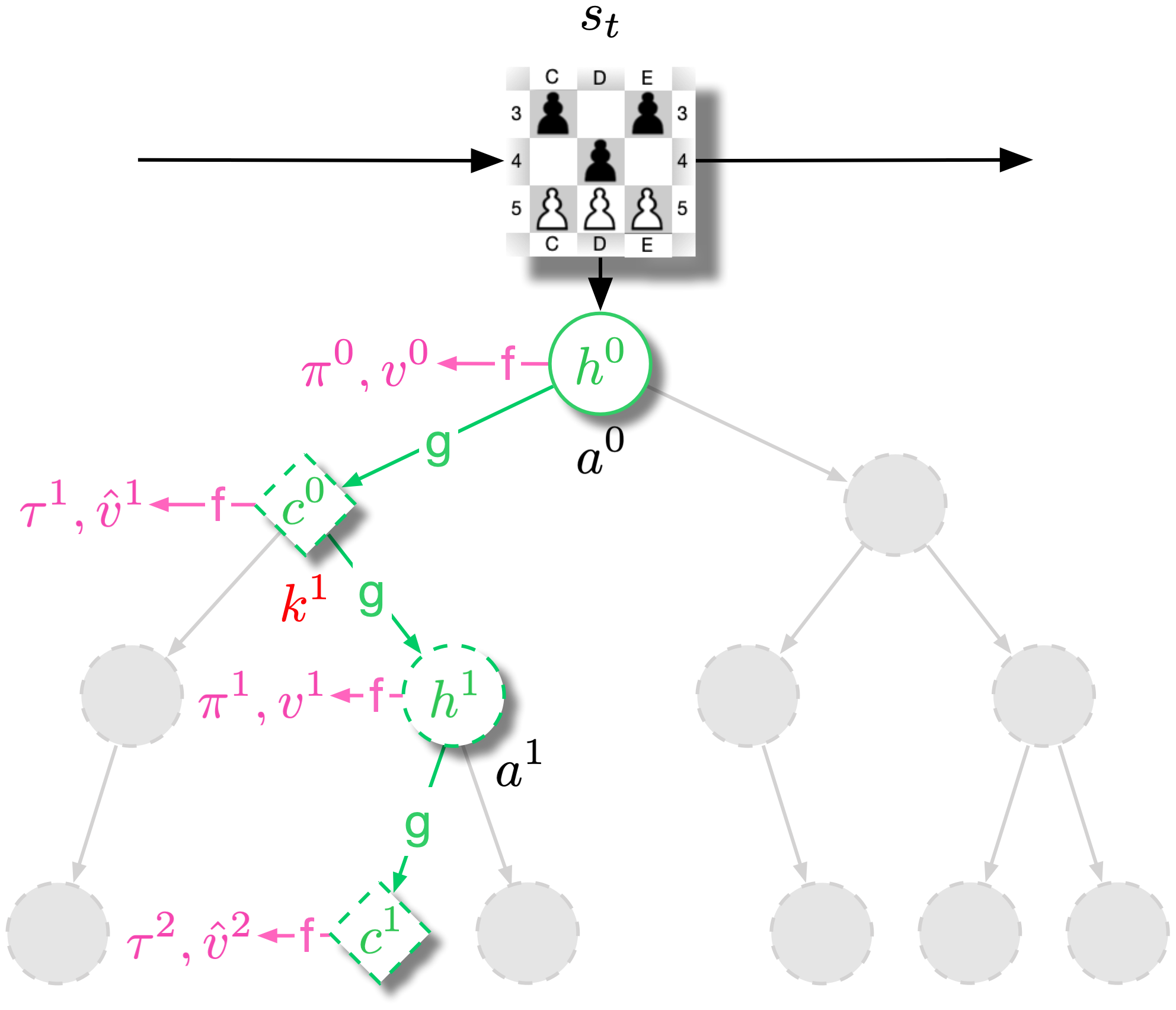}
         \caption{MCTS branches over both actions and state latent variables.}
         \label{fig:alphazero_path}
     \end{subfigure}
     
    \caption{
    The main components of the proposed agent.
    }
    \vskip -0.3in
    \label{fig:summary_figure}
    \end{center}
\end{figure}

Our VQ model is trained with a two-stage training process\footnote{Training the full model end-to-end is a promising future research direction.}. We first train the state VQVAE model which encodes the observations into discrete latent variables (\autoref{fig:state-vqvae}). Then we train the transition model using the discrete latent variables learned by the state VQVAE model (\autoref{fig:transition-model}).

\paragraph{Notation} We use $s_t$, $a_t$, and $r_t$ to denote the state at time $t$, the action following state $s_t$, and the resulting reward, respectively. An episode is a sequence of interleaved states, actions, and rewards ($s_1, a_1, ..., s_t, a_t, ..., s_T$).

\paragraph{State VQVAE}
The purpose of the state VQVAE is to encode a sequence of states and actions into a sequence of discrete latent variables and actions that can reconstruct back the original sequence. This is done by learning a conditional VQVAE encoder-decoder pair. The encoder takes in states $s_1, ..., s_t, s_{t+1}$ and actions $a_1, ..., a_{t}$, and produces a discrete latent variable $k_{t+1} = f_{enc}(s_{1:t+1}, a_{1:t})$. The decoder takes in the discrete latent variable and the states and actions until time $t$ and reconstructs the state at time $t+1$, i.e. $\hat{s}_{t+1} = f_{dec}(s_{1:t}, a_{1:t}, k_{t+1})$. Thus, $k_{t+1}$ represents the additional information in the state $s_{t+1}$ given the previous states and actions.

The state VQVAE is trained using the VQVAE technique introduced in \citet{van2017neural} (reviewed in Section~\ref{sec:vqvae}). The cardinality of the discrete latent variable is a design choice. Larger values are more expressive but potentially expensive at search time. Lower values would lead to lossy compression but could potentially yield more abstract representations. We show the effect of cardinality size on reconstruction quality in the supplementary material.
\autoref{fig:vq_state_model} depicts the state VQVAE.

\begin{figure}[t]
    \centering
    \includegraphics[width=0.99\columnwidth]{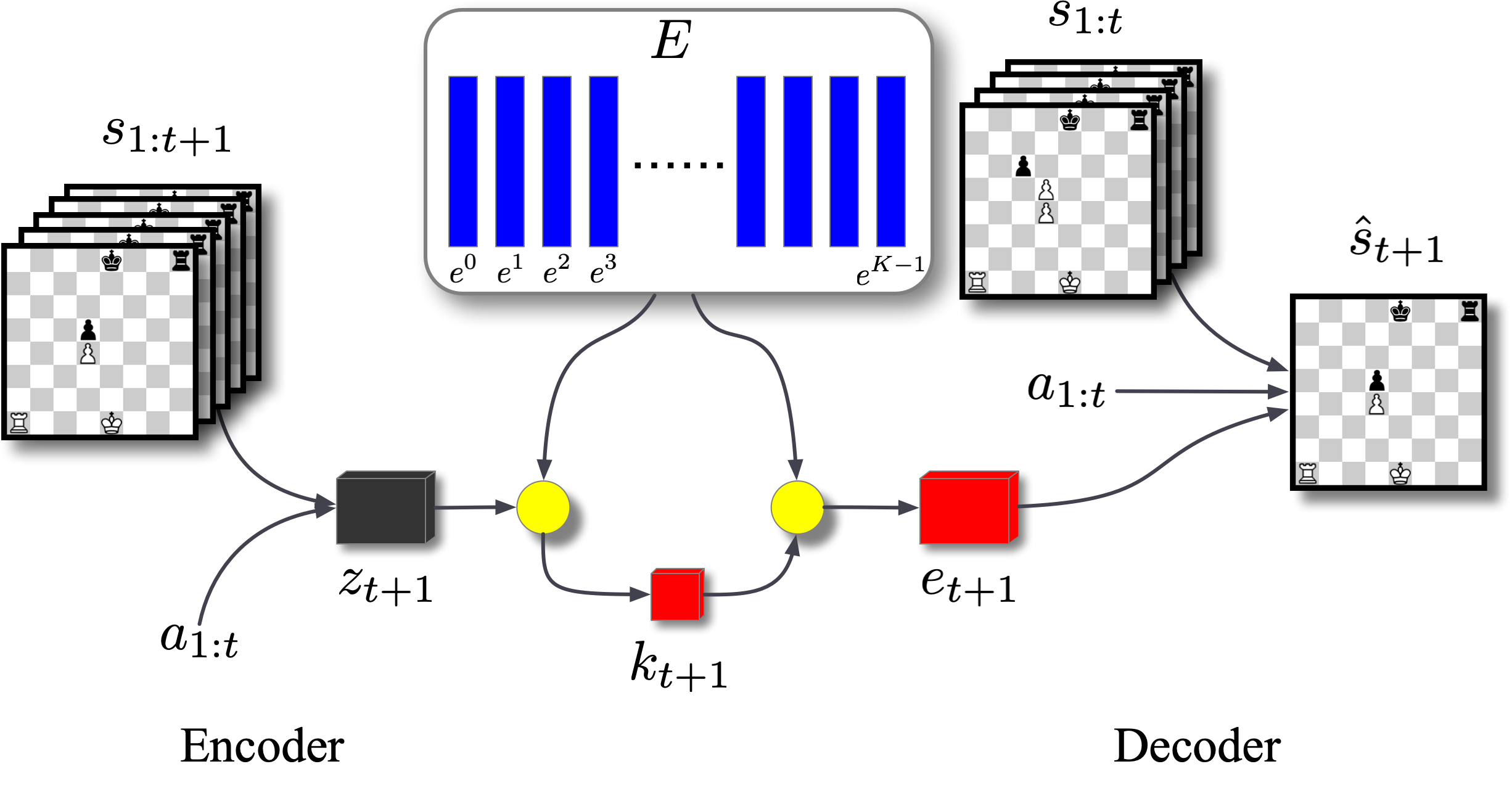}
    \caption{\textbf{Complete encoder/decoder architecture of the state VQVAE.} Encoder compresses $s_{1:t+1}$ and $a_{1:t}$ to a continuous latent $z_{t+1}$. The quantization layer returns the nearest code $e_{t+1}$, as well as the corresponding index $k_{t+1}$, in its codebook $E$. Decoder uses $s_{1:t}$, $a_{1:t}$ and the code $e_{t+1}=E[k_{t+1}]$ to reconstruct $s_{t+1}$.}
    \label{fig:vq_state_model}
\end{figure}

\paragraph{Transition Model}

We obtain latent variable $k_t$ from state $s_t$ using the state VQVAE. We construct a \emph{planning path} which comprises of a state followed by a sequence of interleaved actions and latent variables until a maximum depth $M$ is reached, i.e. $s, a^0, k^{1}, a^{1} \dots, a^{M-1}, k^{M}$. Thus, instead of planning over only the agent's actions, this allows us to also plan over the outcomes of those actions. 

Similar to the environment model of MuZero, our transition model predicts reward and value function at every step. Unlike MuZero, our model not only has a policy head $\pi$ but also a discrete latent code head $\tau$. We alternate the prediction of action and discrete latent code along the sequence. Note that the value prediction after the discrete latent code corresponds to an estimate of the state value functions, while value prediction after the action corresponds to an estimation of the Q-function.

To train all the components, and again following MuZero, we use teacher forcing of trajectories generated by a behavior policy (human experts or other agents in all our experiments) based on the observed states $s$, actions $a$, and latent variables $k$. The total loss combining all the prediction losses is

\begin{align*}
    & \frac{1}{M} \sum_{m = 0}^{M-1} \text{CE}(a^m, \pi(h^{2m})) + \frac{1}{M} \sum_{m=1}^{M} \text{CE}(k^m, \tau(h^{2m-1}))  \nonumber \\  + & \frac{\alpha}{2M} \sum_{m=0}^{2M} \mathcal{L}^v(v_{target}^m, v(h^m)) + \frac{\beta}{2M} \sum_{m=0}^{2M} \mathcal{L}^{r}(r_{env}^m, r(h^m)),
\end{align*}

where $\text{CE}$ is the cross entropy loss. The total loss is similar to MuZero loss. The main difference is that we also predict latent variables at every odd timestep.

Throughout this Section, we explain our VQM-MCTS based on the VQHybrid \emph{planning path} (\autoref{fig:vq_hybrid}). However, VQHybrid is not the only choice available. Depending on the information encoded by the state VQVAE, the \emph{planning path} can be structured differently. 

VQPure (\autoref{fig:vq_pure}) requires a factorized state VQVAE which provides two sets of latent space: one for environment stochasticity, same as in VQHybrid, and the other for the agent's actions. VQPure allows the transition model to decouple from the action space and unroll purely in discrete latent spaces. In contrast, VQHybrid interleaves between state latent variables and actions. For VQJumpy (\autoref{fig:vq_jumpy}), the state VQVAE makes a jumpy prediction of the state $s_{t+m}$ instead of predicting the immediate state $s_{t+1}$.

This enables ``temporal abstractions'' during planning and provides a principled way to unlock planning in a stochastic environment and decoupling the action space tied to environment both in branching and in time. Although we provide some result for VQPure in our chess experiment, these two \emph{planning path} alternatives remain largely unexplored and left for future works.

\subsection{Monte Carlo Tree Search with VQ Planning Model}
\label{sec:vq_mcts}
In order to use the VQ planning model, we modify \emph{Monte Carlo Tree Search} (MCTS) algorithm to incorporate the VQ ``environment action''. The main difference with the MCTS used in MuZero (reviewed in Section ~\ref{sec:mcts}) is that, instead of predicting agent (or opponent) actions, our MCTS also predicts next discrete latent variables $k$ given the past. Unlike classical MCTS which anchors in the real action space and imposes an explicit turn-based ordering for multi-agent environment, our MCTS leverages the abstract discrete latent space induced by the state VQVAE. 

The search tree in MCTS consists of two types of nodes: action node and stochastic node. During selection stage, MCTS descends the search tree from the root node: for action nodes, Equation \ref{eq:mcts_action} is used to select an action; for stochastic nodes, $$\argmax_k \left[\hat{Q}(s, k) + P(k|s) U(s, k)\right]$$ is used to select a discrete latent code. We obtain $U(s, k)$ by replacing $a$ with $k$ in $U(s, a)$ from Equation \ref{eq:mcts_action}. $P(k|s)$ is computed with the learned policy $\tau$ of the discrete latent code. $\hat{Q}(s, k)$ can be $0$ for a neutral environment, $Q(s, k)$ if the environment is known to be cooperative or $-Q(s, k)$ if the environment is known to be adversarial. When $\hat{Q}(s, k)=0$, our algorithm is similar to Expectimax search \citep{Michie1966GAMEPLAYINGAG,russell2016artificial} where the expectation of children's Q-value is computed at the stochastic nodes. 
In zero-sum games like Chess and Go, we can use the adversarial setting. As we will see in Section \ref{sec:chess}, adding this prior knowledge improves agent performance.

%% file: experiments.tex
\section{Experiments}

\begin{figure*}[t]
\centering
\begin{minipage}[t]{0.45\textwidth}
        \includegraphics[width=\textwidth]{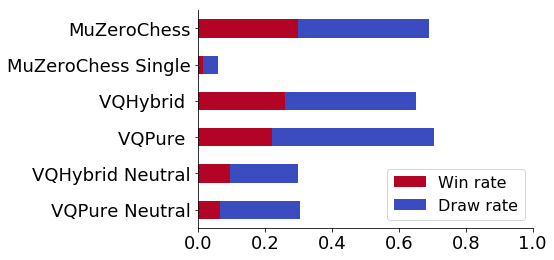}
        \caption{Performance of agents playing against Stockfish 10 skill level 15.}
        \label{fig:chess-results}
\end{minipage}\hfill
    \centering
    \begin{minipage}[t]{0.45\textwidth}
    \includegraphics[width=\textwidth]{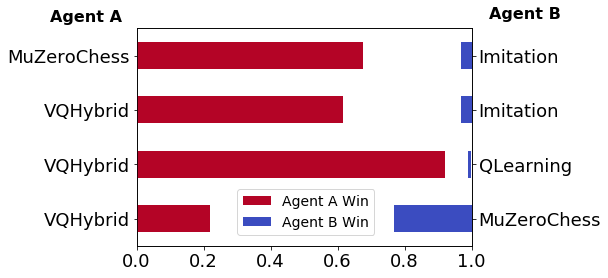}
    \caption{Performance of agents playing against each other.}
    \label{fig:vq-vs-baselines}
    \end{minipage}
\end{figure*}

Our experiments aim at demonstrating all the key capabilities of the VQ planning model: handling stochasticity, scaling to large visual observations and being able to generate long rollouts, all without any performance sacrifices when applied to environments where MCTS has shown state-of-the-art performance.

We conducted two sets of experiments: in Section \ref{sec:chess}, we use chess as a test-bed to show that we can drop some of the assumptions and domain knowledge made in prior work, whilst still achieving state-of-the-art performance; in Section \ref{sec:dmlab}, with a rich 3D environment (DeepMind Lab), we probe the ability of the model in handling large visual observations in partially observed environment and producing high quality rollouts without degradation.

\subsection{Chess}
\label{sec:chess}
Chess is an ancient game widely studied in artificial intelligence \citep{shannon1950xxii}. Although state transitions in chess are deterministic, the presence of the opponent makes the process stochastic from the agent's perspective, when the opponent is considered part of the environment. 
\subsubsection{Datasets}
To evaluate our approach, we follow the two-stage training of VQM and use MCTS evaluation steps illustrated in Section \ref{sec:approach}. We use the offline reinforcement learning setup by training the models with a fix dataset. We use a combination of Million Base dataset (2.5 million games) and FICS Elo >2000 dataset (960k games)\footnote{https://www.ficsgames.org/download.html}. The validation set consists of 45k games from FICS Elo>2000 from 2017. The histogram of player ratings in the datasets is reported in the supplementary material.
\subsubsection{Model Architectures}
The state VQVAE uses feed-forward convolutional encoder and decoder, along with a quantization layer in the bottleneck. The quantization layer has 2 codebooks, each of them has 128 codes of 64 dimensions. The final discrete latent is formed by concatenating the 2 codes, forming a 2-hot encoding vector.

The transition model consists of a recurrent convolutional model which either takes an action or a discrete latent code as input at each unroll step. The model predicts the policies over action and discrete latent code, as well as the value function. We use Monte Carlo return of the game as the target value for training.
\subsubsection{Evaluations}

The performance of our agent is evaluated by playing against:
\begin{enumerate}
    \item Stockfish version 10 \citep{stockfish} (44 threads, 32G hash size and 15s per move);
    \item Q-value agent: Action with the highest Q value is picked. The Q value is computed by unrolling the model for one step and using the learned value function to estimate the value of the next state.
    \item Imitation agent: Agent chooses the most probable action according to the learned policy. 
    \item MuZeroChess agent: MuZero with the same backbone architecture as VQ planning model.
\end{enumerate} 
Each agent is evaluated for 200 games playing as white and black respectively for 100 games. We present results for both the worst case and neutral scenario of VQM-MCTS with VQHybrid and VQPure planning path. 

\subsubsection{Results}

\begin{figure}[!ht]
     \centering
        
        \includegraphics[width=\columnwidth]{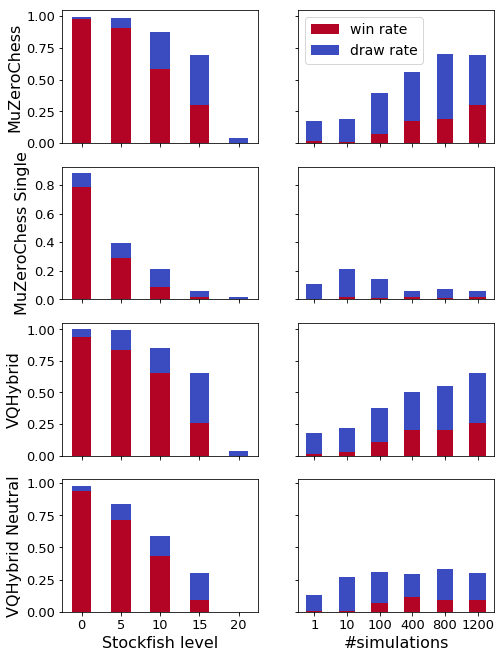}
    
  \caption{\textbf{Agent performance as function of Stockfish strength and simulation budget.} 
  Left column shows win and draw rates of the agent evaluated by playing against different levels of Stockfish 10 with a fixed simulation budget of 1200 per move. Right column shows the impact of simulation budget on agent performance playing against level 15. 
  }
  \label{fig:compare-sim-level}
    \vskip -0.1in  
\end{figure}

\autoref{fig:chess-results} reports our main results. Performance of Single-player MuZeroChess (which doesn't observe opponent actions) is considerably worse than two-player MuZeroChess. In addition, \autoref{fig:compare-sim-level} shows that using more MCTS simulations hurts single-player MuZeroChess performance, because the model is not correctly accounting for the stochasticity introduced by the unobserved opponent's actions. Both VQHybrid and VQPure agents with worst case chance nodes are able to recover the performance to the same level as the two-player MuZeroChess agent. We then further remove the assumption of the adversarial environment by using neutral case chance nodes during MCTS. The resulting agents, VQHybrid Neutral and VQPure Neutral, don't perform as well as VQHybrid and VQPure. This shows that prior knowledge of the environment can indeed help the agent perform better. 

We also note that when assuming the environment is neutral, increasing the simulation budget doesn't seem to improve the performance (\autoref{fig:compare-sim-level}). This is because MCTS searches over the expected behavior of players from a broad range of Elo ratings as human data is used for training. This expectation could deviate significantly from the Stockfish agent, especially for higher skill levels. The full results for the agents can be found in the supplementary material. 

In \autoref{fig:vq-vs-baselines}, we report win and draw rates of a VQHybrid agent playing against other baseline agents. The results confirm that the VQ agent's performance is on par with two-player MuZeroChess: VQHybrid and MuZeroChess achieve very similar performance playing against the same imitation agent; when playing directly against each other, VQHybrid and MuZeroChess reach similar win rate. 

\subsection{DeepMind Lab}
\label{sec:dmlab}

\begin{figure*}[ht]
    \centering
    \includegraphics[width=0.99 \textwidth]{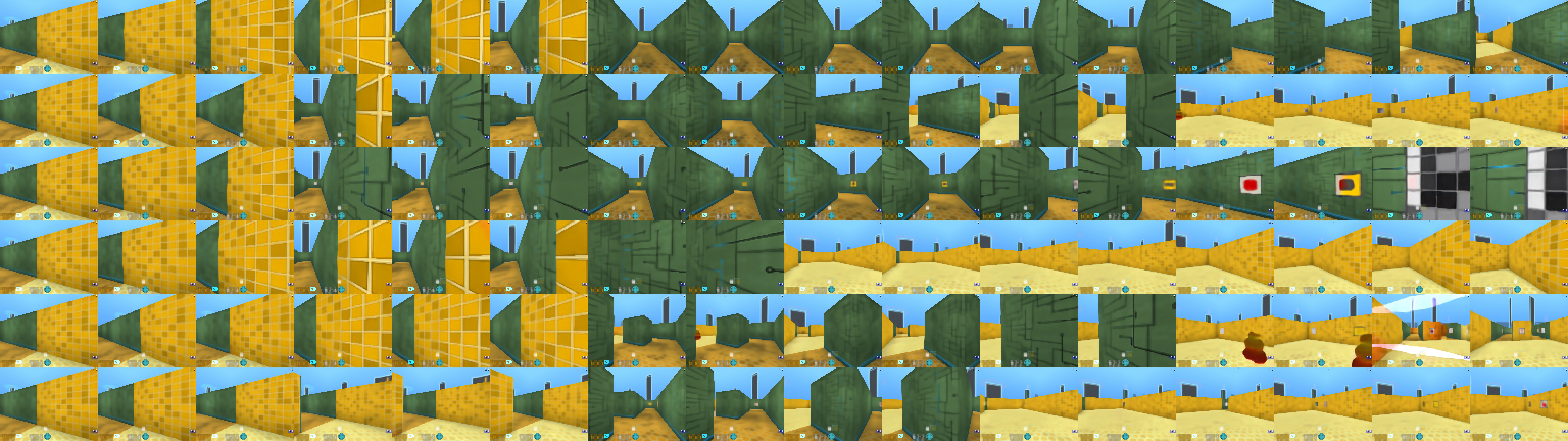}
    \caption{Different random rollouts from the same prefix video in \textit{DeepMind Lab}. The first row is a ground-truth episode from the validation set. Each of the following rows show a sample from the model conditioned on the same 16 starting frames (not shown in the figure). Videos of more samples can be seen in \url{https://sites.google.com/view/vqmodels/home}.}
    \label{fig:dmlab-results}
    
\end{figure*}

\textit{DeepMind Lab} is a first-person 3D environment that features large and complex visual observations, and stochasticity from procedural level generation and partial observability. Due to these properties, it makes for a good environment to test the scalability of our VQ planning model.

\subsubsection{Dataset}
We used an A2C agent \citep{mnih2016asynchronous} to collect a dataset of $101,325,000$ episodes from the \texttt{explore\_rat\_goal\_locations\_small} level in \textit{DeepMind Lab}, $675,500$ of which is held out as the test set. Each episode has 128 timesteps in which the agent is randomly spawned in a maze-like environment, which it observes from first-person view. The agent is rewarded when it captures one of the apples that are randomly placed in the environment, at which point it is transported to a new random location in the map. The collection of episodes is started with a randomly initialized agent and is continued as the training of the agent progresses and it learns about the environment. The collected dataset thus comprises a variety of episodes corresponding to different experience level of the A2C agent.

\subsubsection{Model Architecture and Training}
In addition to the approach described in Section~\ref{sec:approach}, we add a frame-level VQVAE training stage at the start, which uses feed-forward convolutional encoder and decoder trained to map observed frames to a frame-level latent space. The codebook has 512 codes of 64 dimensions. 
Then as discussed in Section \ref{sec:approach}, in the second stage, we train the state VQVAE on top of the frame-level VQ representations. This model captures the temporal dynamics of trajectories in a second latent layer, consisting of a stack of 32 separate latent variables at each timestep, each with their separate codebook comprising of 512 codes of 64 dimensions. The architecture for this component consists of a convolutional encoder torso, an encoder LSTM, a quantization layer,  a decoder LSTM and finally a convolutional decoder head that maps the transition discrete latents back to the frame-level VQ space. Finally in the third stage, we fit the transition model with a hybrid ``planning path'' using a deep, causal Transformer \citep{vaswani2017attention}.


\par
To generate new samples from the model, we first sample state latent variables from the prior network. These are then fed to the state VQVAE decoder for mapping back to the frame-level discrete latent space. The resulting frame-level codes are mapped to the pixel space by the frame-level VQVAE decoder.

\subsubsection{Baselines}
We compare our proposed approach based on VQM with several baselines. As the simplest baseline, we train a deterministic next-frame prediction model with an LSTM architecture closely mimicking the state VQVAE architecture except for the quantization layer. The network is trained to reconstruct each frame given the preceding frames with mean-squared-error (MSE). Additionally, we train several sequential continuous VAE baselines with different posterior and prior configurations to compare our discrete approach with continuous latent variables.
We use GECO \citep{rezende2018geco} to mitigate the well-known challenges of training variational autoencoders with flexible decoder and prior models. In particular, we assign a target average distortion tolerance for the decoder and minimize, using the Lagrange multiplier method, the KL-divergence of the posterior to the prior subject to this distortion constraint, as described in \citet{rezende2018geco}. We choose two different distortion levels of 25dB and 33dB PSNR. The former is based on our experiments with the deterministic LSTM predictor (the first baseline described above), which achieves reconstruction PSNR of about 24dB, and the latter is set slightly higher than the reconstruction PSNR of our frame-level VQVAE decoder at 32dB.

\subsubsection{Evaluation Metric}
For every trajectory in the test set we take $k(=1000)$ sample episodes from the model using the initial prefix of $T_0(=16)$ frames from each ground-truth trajectory. For each ground-truth trajectory, we find the sample with minimum (cumulative) reconstruction error (measured as Mean Squared Error or MSE) from the end of the prefix up to a target timestep $t(=128)$, and average the error for this \emph{best-match} sample over the test trajectories. We refer to this metric as \emph{Mean Best Reconstruction Error} or \texttt{MBRE} defined as 
\begin{equation*}
  \texttt{MBRE}(S, G, T_0, T) = \frac{1}{|G|} \sum_{i=1}^{|G|}\min_{s \in S_i} \sum_{t=T_0}^{T} \|s(t) - G_{i}(t)\|^2 , \label{eq:mbre}
\end{equation*}
where $G$ is a set of ground-truth trajectories, and $S$ is a set of sampled episodes--$k$ episodes sampled using the initial $T_0$ frames of each ground-truth episode $G_i$ as prefix. We compare sampled episodes with their corresponding ground truth episode at the target frame $T$. An ideal model that generalizes well to the test set would place a non-negligible probability mass on the ground-truth trajectory, and thus an episode close to the ground-truth should be sampled given a sufficiently large number of trials.

\subsubsection{Results}
\begin{figure}[ht]
    \centering
    
    \begin{subfigure}[ht]{0.85\columnwidth}
         \centering
         \includegraphics[width=\columnwidth]{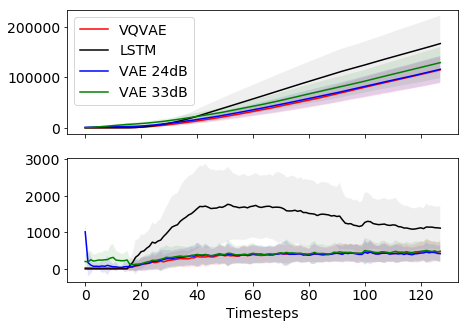}
         \label{fig:dmlab-result-cumulative}
     \end{subfigure}     
     
    \vskip -0.2 in
    \caption{\textit{Mean Best Reconstruction Error} (MBRE) for our proposed VQVAE approach and baselines. X-axis is length of trajectory. Top: computed over frames up to the target frame, Bottom: computed for the target frame only.
    }
    \vskip -0.2in
    \label{fig:dmlab-metric}
\end{figure}

Rollouts generated by the model can be seen in \autoref{fig:dmlab-results}. \autoref{fig:dmlab-metric} shows the results of evaluating MBRE for our proposed approach against the aforementioned baselines. The LSTM baseline performs significantly worse than all latent variable models. This is expected because the generative model is not expressive enough for stochastic and multi-modal data, and as a result predicts the mean of all possible outcomes, which results in poor best matches against ground-truth. 
Comparisons with the two sequential VAE baselines demonstrate the trade-off between the reconstruction quality of a model and its predictive performance as measured by MBRE. Furthermore, it shows that our VQM approach is able to achieve competitive performance with respect to both reconstruction quality and long range predictive performance. 

%% file: discussion.tex
\section{Conclusion}
\label{sec:discussion}
In this work, we propose a solution to generalize model-based planning for stochastic and partially-observed environments. Using discrete autoencoders, we learn discrete abstractions of the state and actions of an environment, which can then be used with discrete planning algorithms such as MCTS \citep{coulom2006efficient}. We demonstrated the efficacy of our approach on both an environment which requires deep tactical planning and a visually complex environment with high-dimensional observations.
Further we successfully applied our method in the offline setting. Our agent learns from a dataset of human chess games and outperforms model-free baselines and performs competitively against offline MuZero and Stockfish Level 15 while being a more general algorithm. We believe the combination of model-based RL and offline RL has potential to unlock a variety of useful applications in the real world.

\section*{Acknowledgements}
We'd like to thank Ivo Danihelka and Nando de Freitas for providing valuable feedback on early drafts of the paper. We'd like to thank Julian Schrittwieser for helping with the MuZero baselines. We'd also like to thank Sander Dielman, David Silver, Yoshua Bengio, Jakub Sygnowski, and Aravind Srinivas for useful discussions and suggestions.

%% file: appendix.tex
\section{Appendix}

\subsection{Chess datasets}
\label{appendix:ratings}
\begin{figure}[!ht]
     \centering
     \begin{subfigure}[h]{0.3\textwidth}
         \centering
         \includegraphics[width=\textwidth]{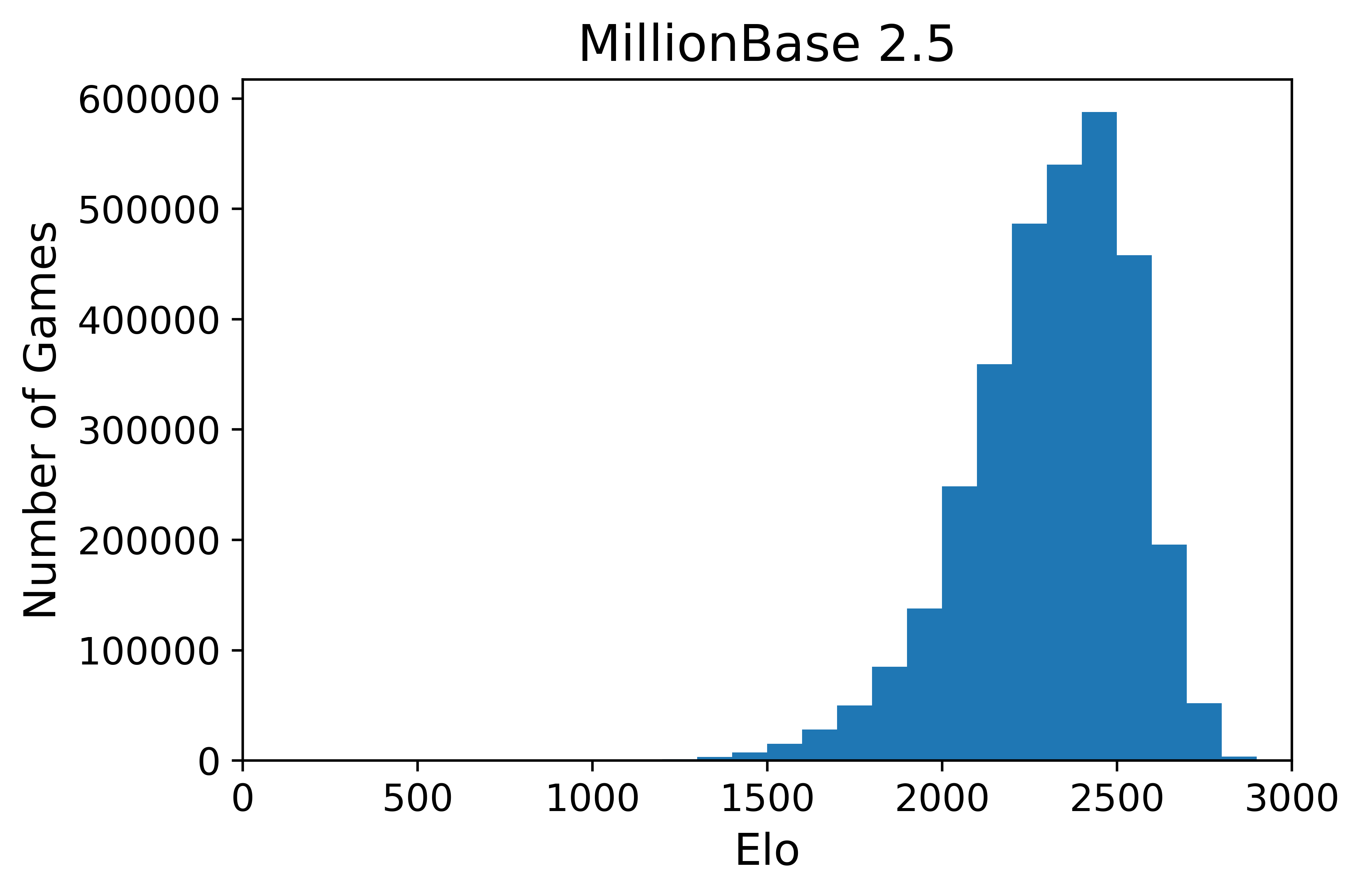}
     \end{subfigure}
     \hfill
     \centering
     \begin{subfigure}[h]{0.3\textwidth}
         \centering
         \includegraphics[width=\textwidth]{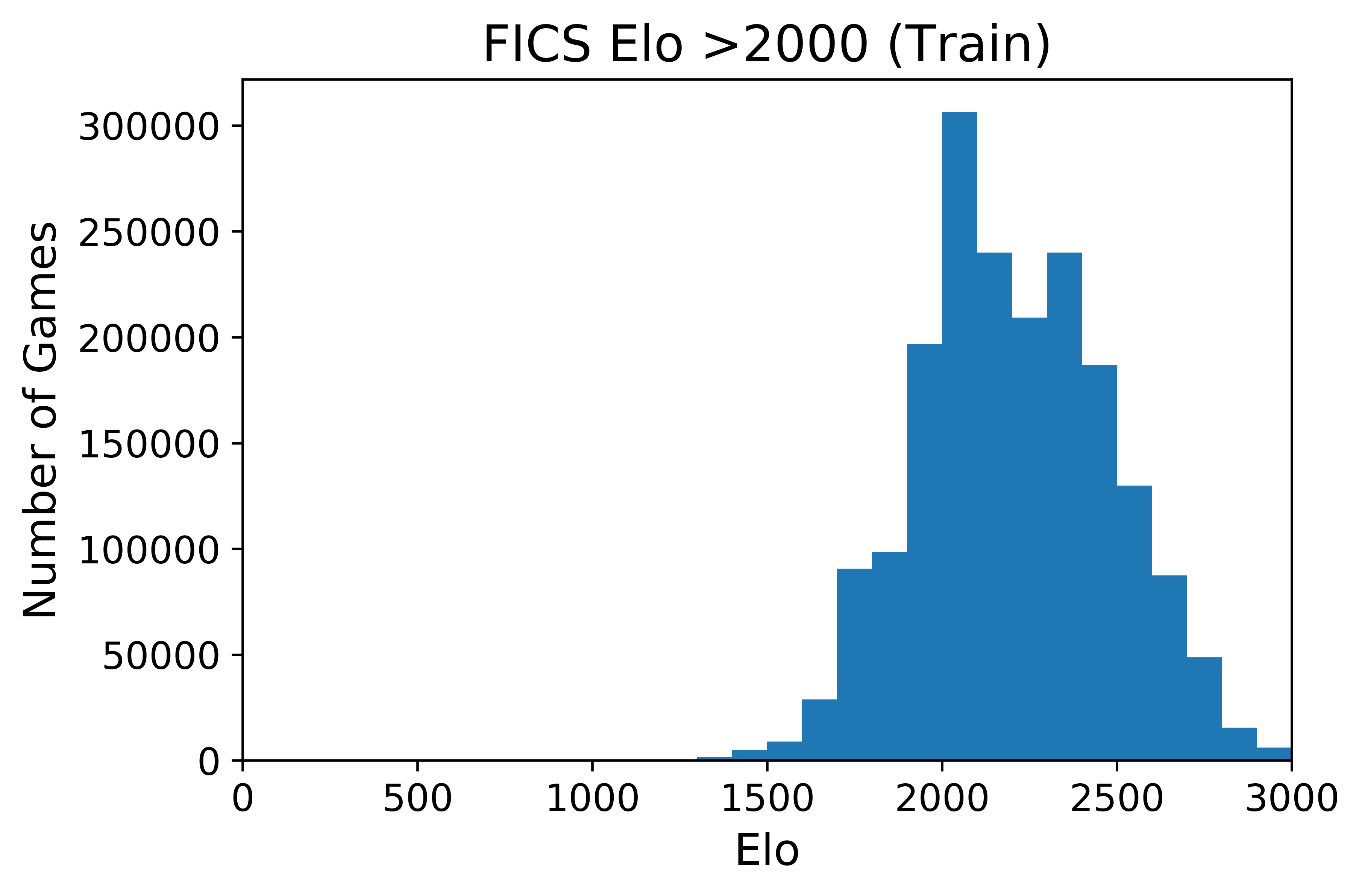}
     \end{subfigure}
\caption{Histogram of Elo rating for players in Million Base dataset and FICS Elo2000 training set.}\label{fig:ratings}
\end{figure}

The histogram of Glicko-2 ratings \citep{glickman2012example} of the players in the training set is shown in \autoref{fig:ratings}.

\subsection{Details for Chess Experiments}
\label{appendix:chess_details}

\subsubsection{Input Representations}
We use symbolic representation for the chess board where each piece is represented as an integer. Following the same protocol as \citet{alphazero}, the action space is represented as $8 \times 8 \times 73$ discrete actions.

\subsubsection{Baseline Models}
\paragraph{MuZeroChess}
The baseline MuZeroChess model is similar to that of the MuZero \citep{schrittwieser2019mastering}. In stead of using a deep convolutional network extracting features from the board state, we simply embed the board state to get the initial state $s$. The dynamics function is consist of 2 convolutional networks $g$ and $f$. It works as follows: set $z^0 = s$ for the initial state; $z^{m+1} = g(z^{m}, a^{m})$, $g$ takes in the embedded action input $a^{m}$ and previous state $z^{m}$ and outputs the next state $z^{m+1}$; $h^{m+1} = f(z^{m+1})$, $f$ produces the final state for prediction heads. We have two prediction heads $v$ and $\pi$ for value and policy respectively. Both $g$ and $f$ are 20-layer residual convolution stacks with 1024 hiddens and 256 bottleneck hiddens. All the heads have a hidden size 256. The action predictor uses kernel size 1 and strides 1 and its output is flattened to make the prediction.



We train with sequence length 10. If the training sequence reaches the end of the game, then after game terminates, we pad random actions and target value, as well as mask the action prediction loss.

We use a batch size of 2048 for training and use Adam optimizer \citep{kingma2014adam} with learning rate $3e^{-4}$ and exponential decay of with decay rate 0.9 and decay steps 100000. To stabilize the training, we apply gradient clipping with maximum clipping value 1. The model is trained for 200k steps.

\paragraph{Q-value and Imitation Agents}
The same MuZeroChess model as described above is used. For Q-value agent, we unroll the dynamics function for 1 step with each legal action. The estimated action value of the Q-value agent is the value prediction of the next state. For imitation agent, we use the policy prediction of the model as the imitation agent's policy.

\subsubsection{VQ Model}
\paragraph{State VQVAE}

The state VQVAE for chess has a encoder and a decoder, which are both 16-layer residual convolution stack with 256 hiddens and 64 bottleneck hiddens. The quantization layer has 2 codebooks, each of them has 64 codes of 256 dimensions. Before feeding into the quantization layer, we apply spatial mean pooling on the features. The reconstruction of the board state is cast as a classification problem where the model predicts the piece type at each position. Therefore, the reconstruction loss is the cross entropy loss. The training batch size is 1024. We use Adam optimizer \citep{kingma2014adam} with learning rate $3e^{-4}$ and exponential decay of with decay rate 0.9 and decay steps 100000. The model is trained for 1.2 million steps.

\paragraph{Transition model}

Our transition model for chess is similar to the model used for MuZeroChess. In addition to the dynamics function $g$ which takes the action as input at each step, we introduce another dynamics function $g^\prime$ which takes the input of discrete latent codes. At alternating steps, instead of using $g$, we obtain the next state by $z^{2m+1} = g^\prime (z^{2m}, k^{2m})$ where $m > 0$. We also introduce an additional prediction head $\tau$ to predict the discrete latent codes. The additional function $g^\prime$ is a 30-layer residual convolution stack with 1024 hiddens and 256 bottleneck hiddens. Unlike for actions, discrete latent codes after termination of the game don't need random padding or masking.

We use the same training setup and optimizer parameters as MuZeroChess. VQHybrid model is trained for 200k steps; VQPure model is trained for 400k steps.

\paragraph{MCTS}

The hyperparameters used for the MCTS are the same for baseline MuZeroChess and VQM-MCTS: $discount=1.0$, UCB parameters $c_{base}=19652.0$ and $c_{init}=1.25$. No temperature is applied on the acting visit count policy. Same as MuZero \citep{schrittwieser2019mastering}, we limit the agent action to all legal moves. MuZeroChess and VQHybrid agents don't have terminal state. VQPure agent reaches terminal state when the same VQ code is taken for 10 consecutive steps in the search tree branch.

For experiments where VQ agent is playing against Q-value, imitation and MuZeroChess agents, we employ similar strategy used for data generation in \citet{schrittwieser2019mastering}. This is because both agents are deterministic, playing among them would result in deterministic games making the evaluation less meaningful. Specifically, instead of selecting the action with the highest visit count at the end of the tree search, we use a stochastic variant. We keep a pool of possible actions which have an action count of at least 1\% the total count $\mathcal{A}=\{a: N(a) > 0.01 N_{max}\}$, and sample an action $a$ according to the probability induced by visit counts $p(a)=\frac{N(a)}{\sum_{a^{'} \in \mathcal{A}} N(a^{'})}$. This stochastic variant of the tree policy is used for the first 30 steps of the game for all MCTS-based agents.

\subsection{Quasirandom Sampling}
\label{appendix:better-sampling}
\begin{figure}[!ht]
  \begin{center}
    \includegraphics[width=0.3\textwidth]{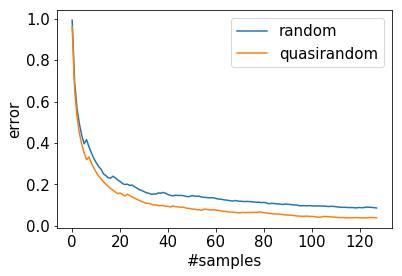}
  \end{center}
  \caption{Quasirandom sampling produces empirical distributions closer to the true distribution than random sampling. The plotted error is the Euclidean distance between the probability distributions. For this analysis, we sampled the probabilities for the Multinomial distrbution from a Dirichlet distribution with $\{\alpha_i\}_{i=1}^N = 1$ where $N=64$.}
\end{figure}

As explained in Section \ref{sec:vq_mcts}, our MCTS implementation takes the following form:
\begin{align*}
\argmax_k \hat{Q}(s, k) + P(k|s) U(s, k) 
\end{align*}
where
\begin{align*}
    U(s, a) &= \frac{\sqrt{N(s)}}{1 +N(s, a)} \left[c_1 + \log \left(\frac{N(s) + c_2 + 1}{c_2}\right) \right],\\
    \hat{Q}(s, k) &= \left\{ \begin{array}{lr}
       Q(s, k) & \text{cooperative} \\
      0 & \text{neutral} \\
      - Q(s, k) & \text{adversarial}
\end{array} \right.
\end{align*}

If we assume the environment to be neutral, the empirical distribution of the selected discrete latent codes at planning time should match the estimated distribution of codes $P(k|s)$. A straightforward way to obtain such a distribution is to sample i.i.d from the estimated distribution. However, in practice, we found that using the above equation with $\hat{Q}(s, k) = 0$ works better. This corresponds to a quasi-random Monte Carlo sample of a multinomial distribution $p_i$ where we simply select the value which has the largest value of $\frac{p_i}{N(i)+1}$, where $N$ is the number of times $i$ has been sampled already.

\subsection{Chess Agents Performance against Stockfish}
\label{appendix:further_chess_results}

\begin{figure*}[ht]
     \centering
     \begin{subfigure}[h]{0.495\textwidth}
         \centering
         \includegraphics[width=\textwidth]{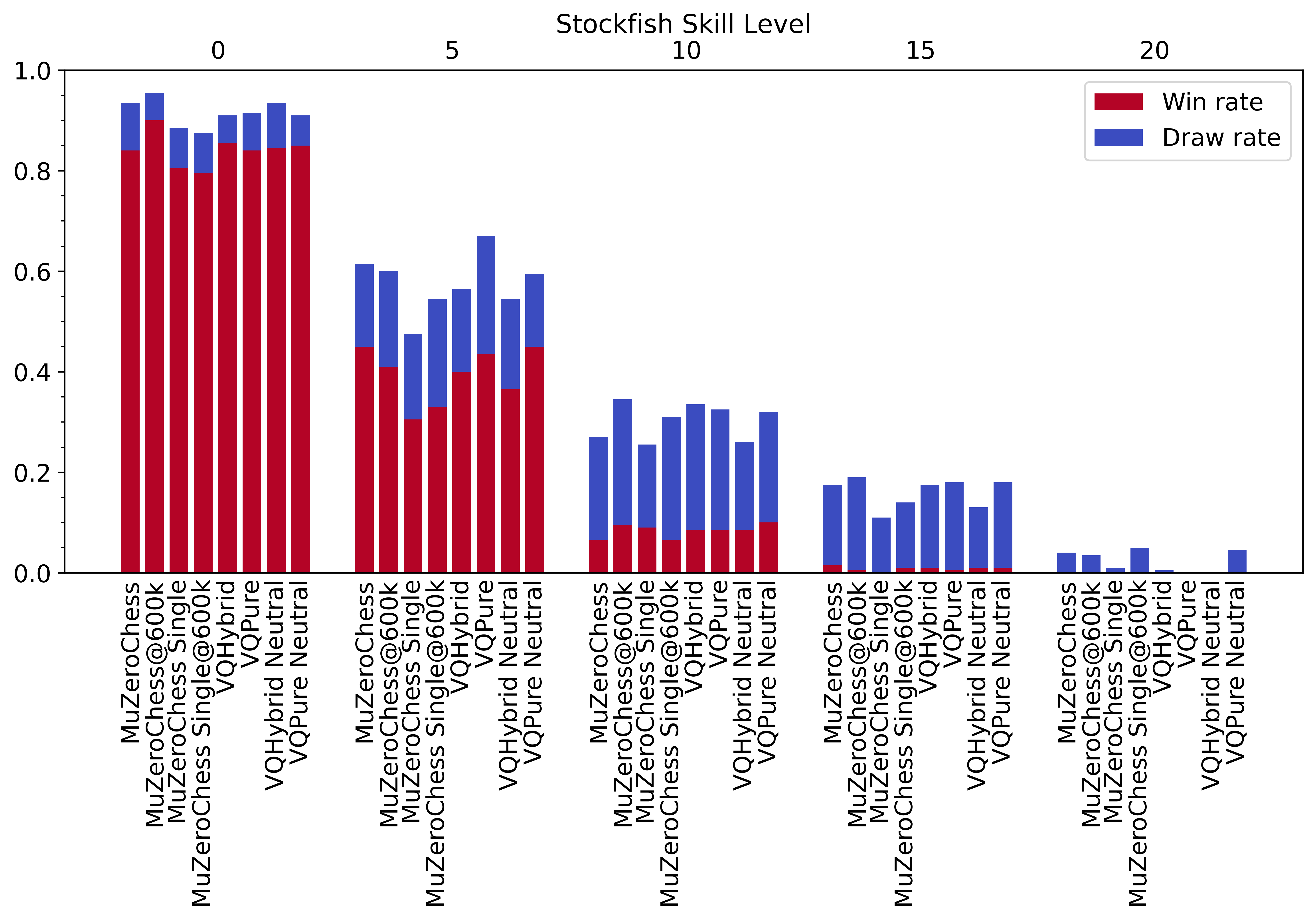}
         \caption{Simulations=1}
         \label{fig:stockfish_sim1}
     \end{subfigure}
     \begin{subfigure}[h]{0.495\textwidth}
         \centering
         \includegraphics[width=\textwidth]{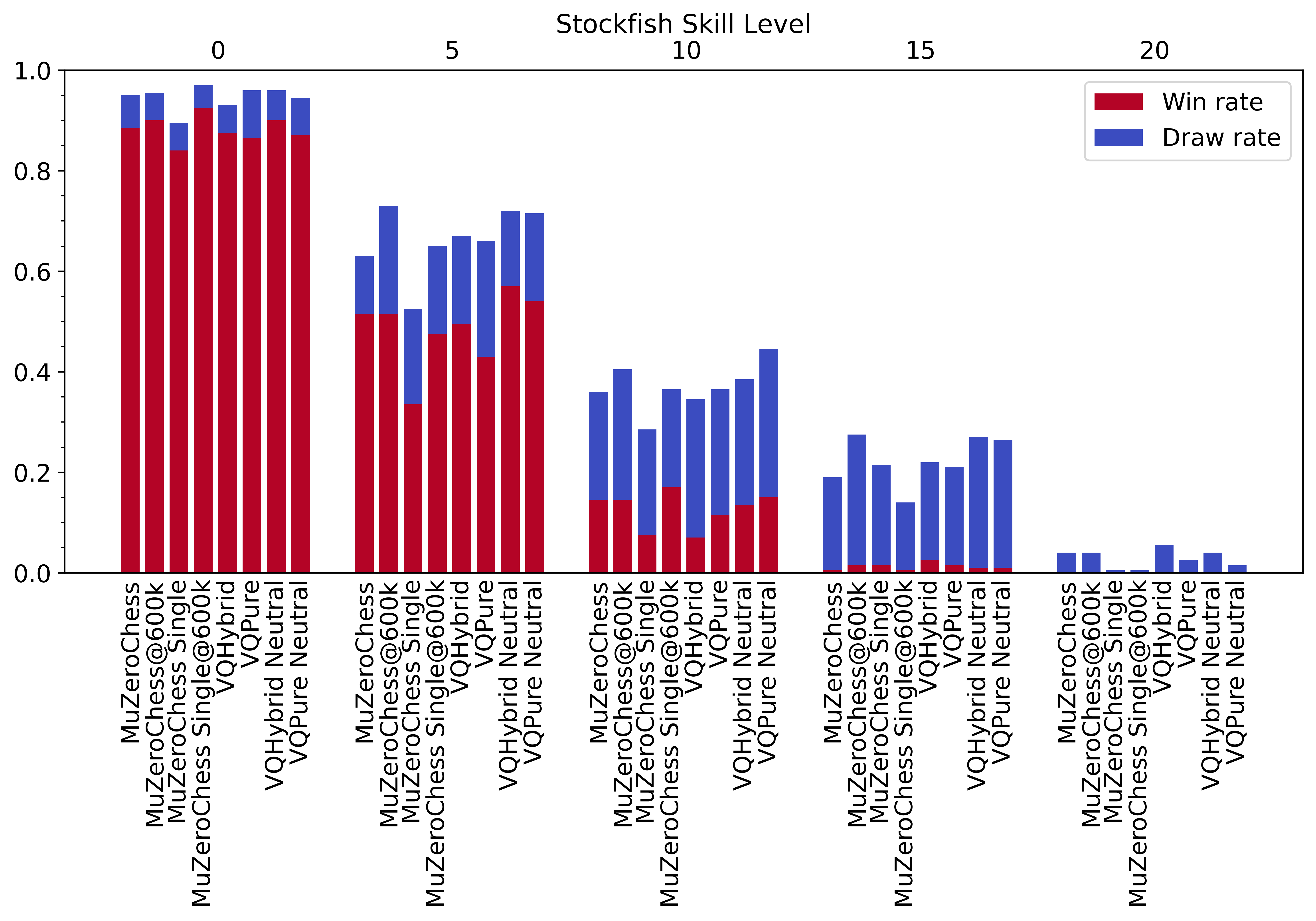}
         \caption{Simulations=10}
         \label{fig:stockfish_sim10}
     \end{subfigure}     
     \begin{subfigure}[h]{0.495\textwidth}
         \centering
         \includegraphics[width=\textwidth]{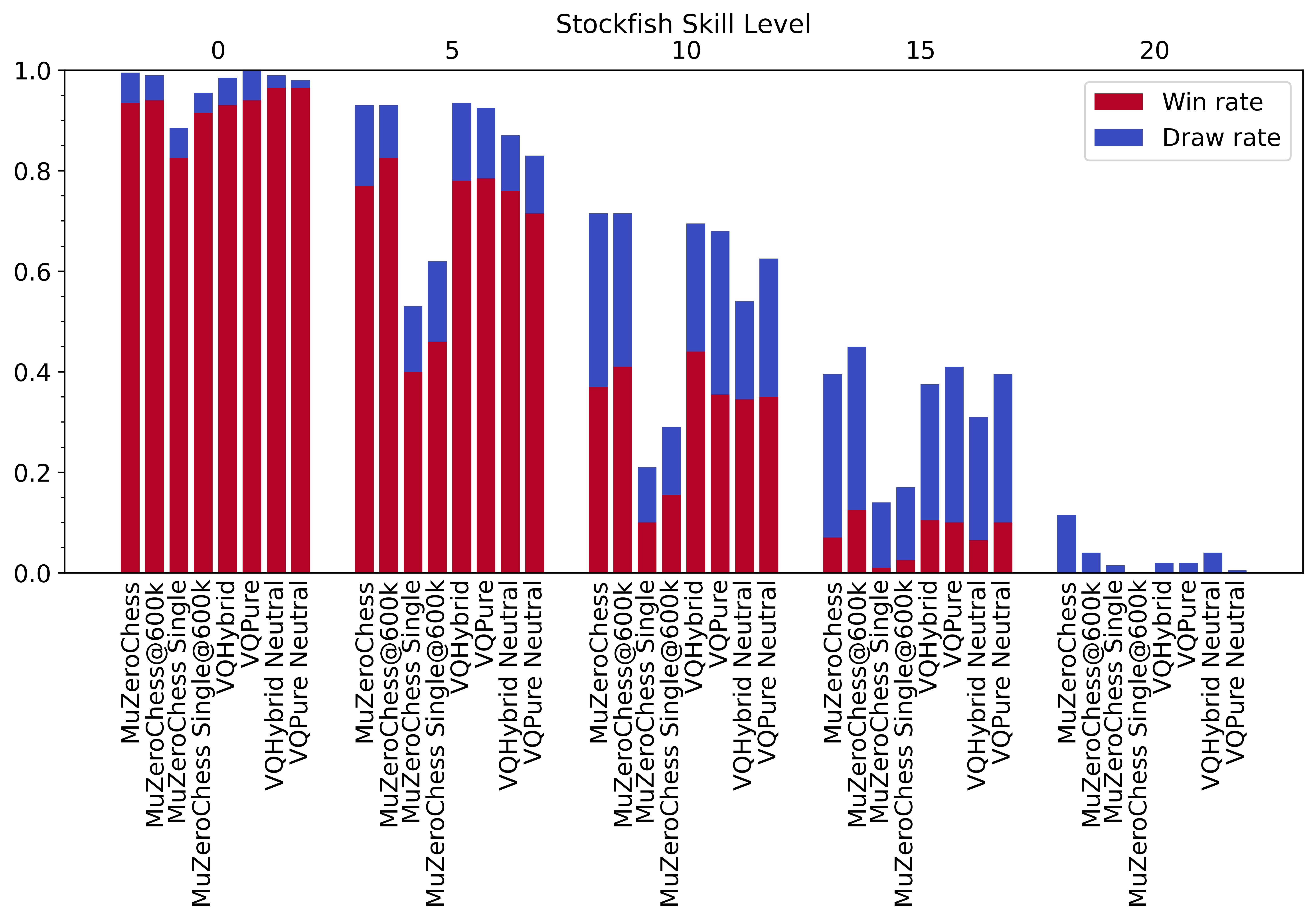}
         \caption{Simulations=100}
         \label{fig:stockfish_sim100}
     \end{subfigure}  
     \begin{subfigure}[h]{0.495\textwidth}
         \centering
         \includegraphics[width=\textwidth]{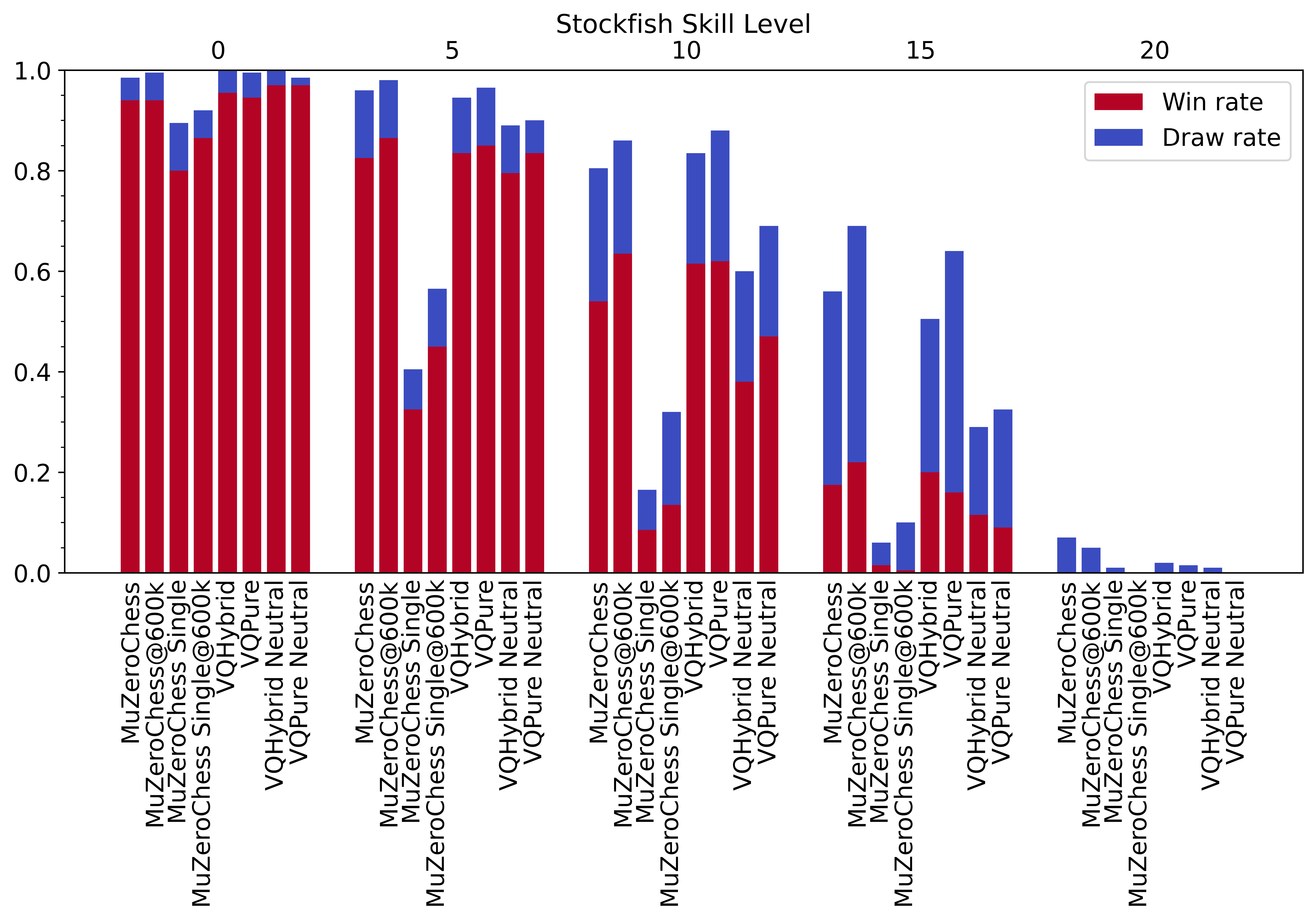}
         \caption{Simulations=400}
         \label{fig:stockfish_sim400}
     \end{subfigure}     
     \begin{subfigure}[h]{0.495\textwidth}
         \centering
         \includegraphics[width=\textwidth]{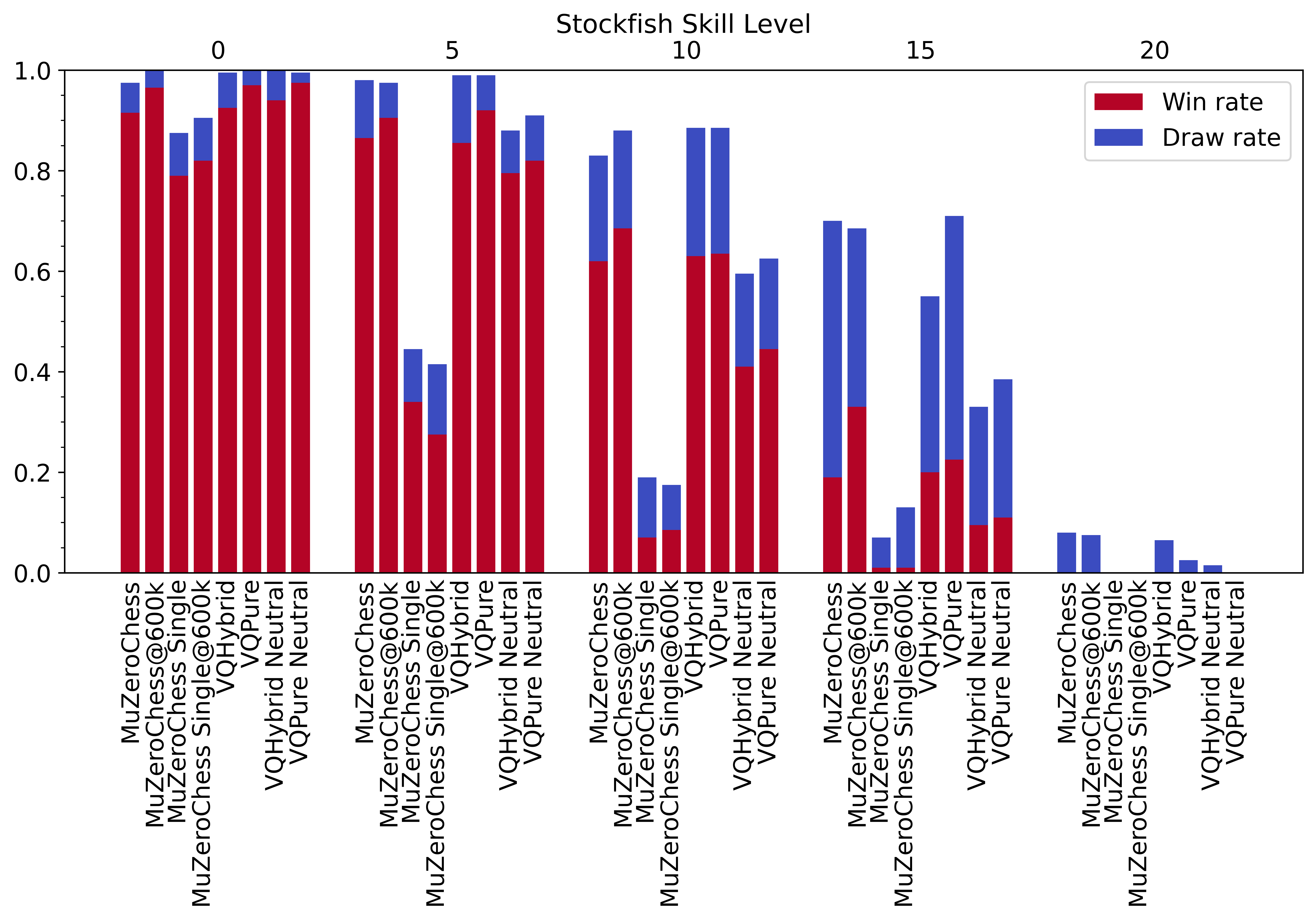}
         \caption{Simulations=800}
         \label{fig:stockfish_sim800}
     \end{subfigure}
     \begin{subfigure}[h]{0.495\textwidth}
         \centering
        \includegraphics[width=\textwidth]{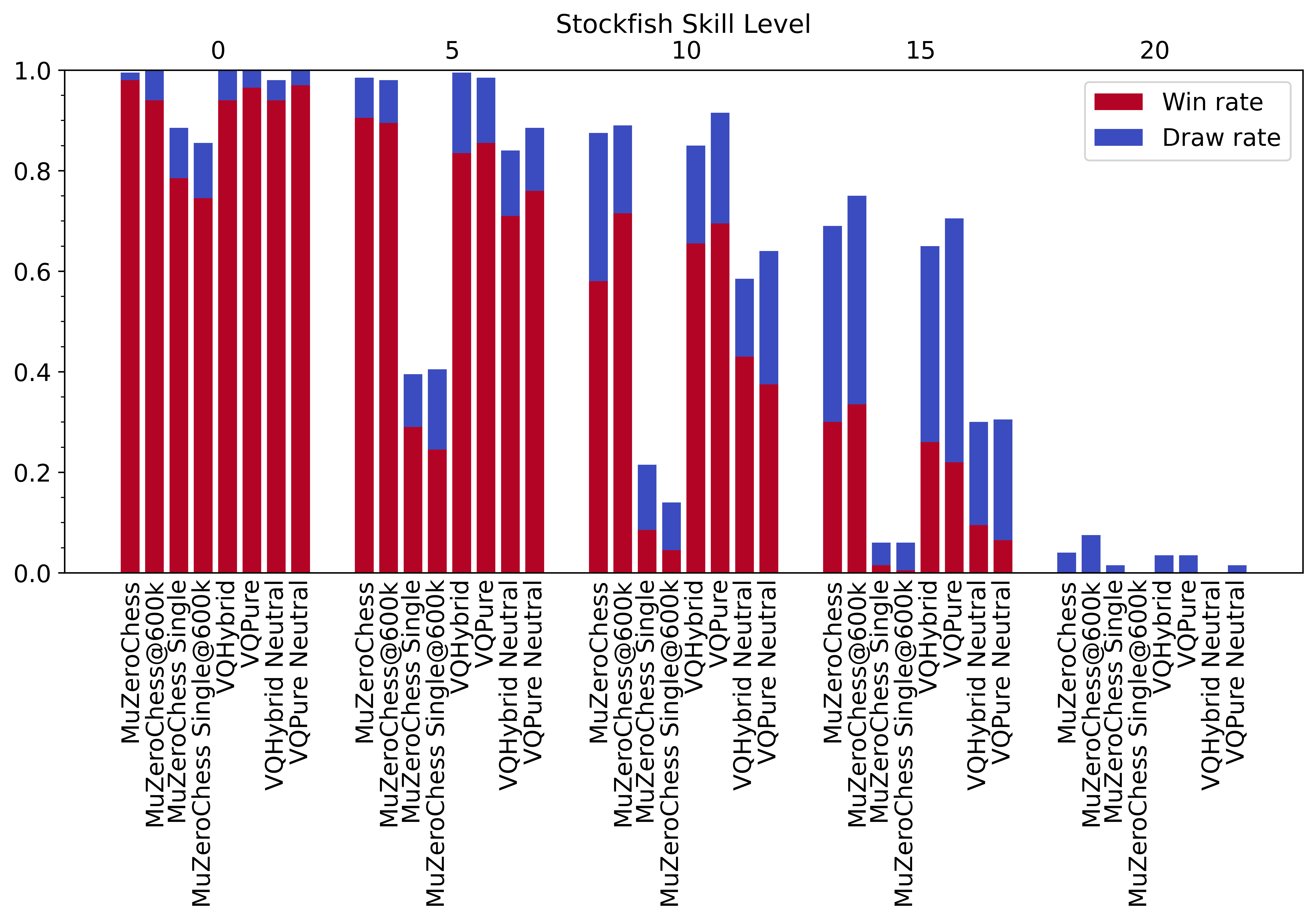}
         \caption{Simulations=1200}
         \label{fig:stockfish_sim1200}
     \end{subfigure}
  \caption{\textbf{Agent performances evaluated against Stockfish 10 with 1, 10, 100, 400, 800 and 1200 simulations per step.} Compare agent performance with different number of simulation budget per move. Reported for agents: two-player MuZeroChess, two-player MuZeroChess@600k, single-player MuZeroChess, single-player MuZeroChess@600k, VQHybrid and VQPure searching over worst case scenario, VQHybrid and VQPure searching over neutral scenario. Stockfish 10 skill levels is varied between 0, 5, 10, 15 and 20 to control the strength of the engine.Red bar shows the win rate; blue bar shows the draw rate of the agents.}
\label{fig:further-chess-results}
\end{figure*}

Because VQ Models have a pre-training phase, a confounding factor for the performance comparison is the amount of computational resources required. We increase the computational resources for MuZeroChess baselines for 3 times. We call the resulting agents MuZeroChess@600k and MuZeroChess Single@600k. This increase is more than that of the pre-training phase since our state VQVAE is trained on a pair of observations whereas the transition model is trained on a sequence of 10 steps. \autoref{fig:further-chess-results} reports the full results of MuZeroChess and VQ agents playing against Stockfish. As \autoref{fig:further-chess-results} shows, with the increase of computational budget, MuZeroChess agent does have a slight performance improvement. However, this is not significant enough to affect the conclusions. Most importantly, MuZeroChess Single agent does not perform better even with significantly more compute. In fact, we see a decrease in win rate across all the Stockfish levels with 1200 simulations.